\documentclass[10pt,journal,twoside]{IEEEtran}
\usepackage{amsmath,amsfonts}
\usepackage{algorithmic}
\usepackage{array}
\usepackage[caption=false,font=normalsize,labelfont=sf,textfont=sf]{subfig}
\usepackage{textcomp}
\usepackage{stfloats}
\usepackage{url}
\usepackage{verbatim}
\usepackage{graphicx}
\def\BibTeX{{\rm B\kern-.05em{\sc i\kern-.025em b}\kern-.08em
    T\kern-.1667em\lower.7ex\hbox{E}\kern-.125emX}}
\usepackage{balance}
\usepackage{graphicx}
\usepackage{cite}
\usepackage{multirow}
\usepackage{url}
\usepackage{colortbl}
\usepackage{amssymb}
\usepackage{mathtools}
\usepackage{bbm}
\usepackage{xcolor}
\usepackage{booktabs}

\definecolor{cvprblue}{rgb}{0.21,0.49,0.74}
\definecolor{red}{rgb}{1.0,0.0,0.0}
\definecolor{urlpink}{RGB}{236,2,141}
\definecolor{tableblue}{RGB}{220, 239, 220}

\definecolor{tablefirst}{RGB}{255, 192, 202}
\definecolor{tablesecond}{RGB}{255, 227, 181}
\definecolor{tablethird}{RGB}{255, 255, 223}

\usepackage[breaklinks,colorlinks=true]{hyperref}

\usepackage[font=small,labelfont=bf,tableposition=top]{caption}

\usepackage{blindtext}
\title{here title}

\let\oldtwocolumn\twocolumn
\renewcommand\twocolumn[1][]{%
    \oldtwocolumn[{#1}{
    \vspace{-10pt}
    \begin{center}
           \includegraphics[width=17cm]{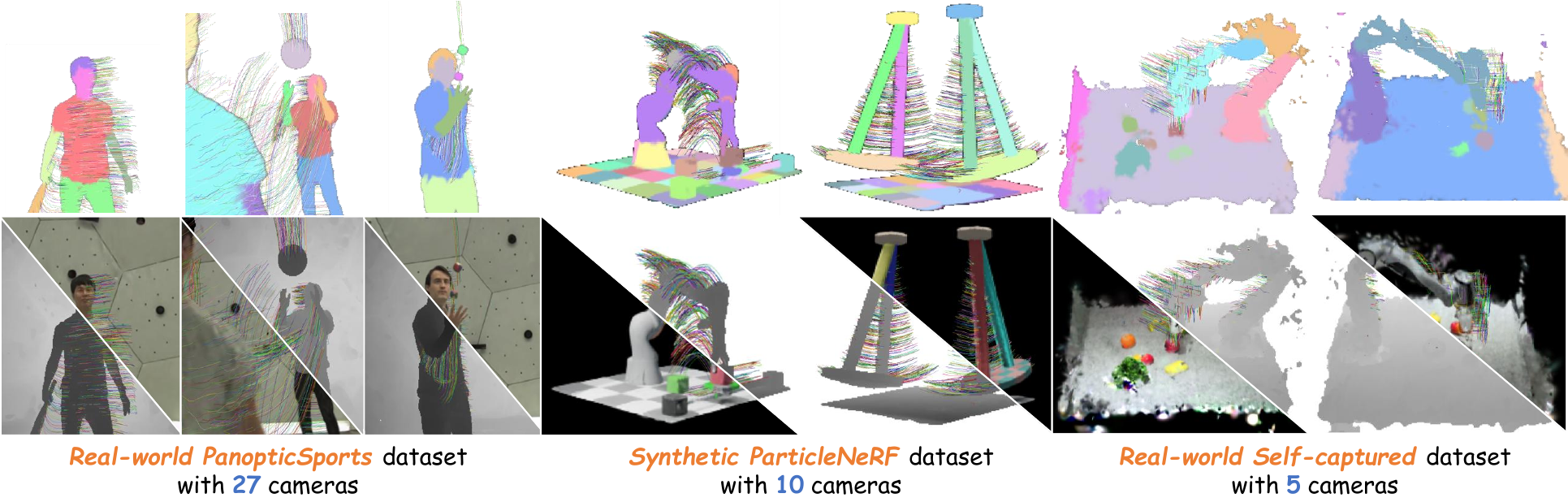}
           \captionof{figure}{{\textbf{Persistent dynamic novel view synthesis (Gaussian parts, colors, and depths) and dense tracking results.}
            The scene is parameterized by a set of 3D Gaussians with evolving centers, rotations, and colors (optional to model changes in shadows), with each Gaussian consistently representing an occupied region in the physical space}. Part colors are randomly assigned for visual distinction. The trajectories display the movement of 4$\%$ of the Gaussians, with their centers tracked across 10 consecutive timestamps. For more video results, please visit PaMoSplat's project website \url{https://pamosplat.github.io/}.}
           \label{fig_teaser}
        \end{center}
    }]
}

\makeatletter
\def\ps@IEEEtitlepagestyle{%
  \def\@oddfoot{%
    \parbox{\textwidth}{\centering\footnotesize
    Copyright © 2026 IEEE. Personal use of this material is permitted. \\
    However, permission to use this material for any other purposes must be obtained 
    from the IEEE by sending an email to pubs-permissions@ieee.org.}%
  }%
  \def\@evenfoot{}%
}
\makeatother

\begin{document}

\markboth{IEEE Transactions on Circuits and Systems for Video Technology}%
{IEEE Transactions on Circuits and Systems for Video Technology}

\title{PaMoSplat: Part-Aware Motion-Guided Gaussian Splatting for Dynamic Scene Reconstruction
\author{ Yinan Deng, Jianyu Dou, Jiahui Wang, Jingyu Zhao, Yi Yang, and Yufeng Yue

\thanks{Manuscript received 31 January 2025; revised 31 May 2025 and 03 December 2025. Date of publication XXXX; date of current version XXXX.
This work is supported by the National Natural Science Foundation of China under Grant 625B2024, 62473050, 92370203, Beijing Natural Science Foundation Undergraduate Research Program QY25270. \textit{(Corresponding authors: Yufeng Yue.)}

Yinan Deng, Jianyu Dou, Jiahui Wang, Jingyu Zhao,  Yi Yang, and Yufeng Yue are with School of Automation, Beijing Institute of Technology, Beijing 100081, China (e-mail: dengyinan@bit.edu.cn; 3120255746@bit.edu.cn; wjh@bit.edu.cn; 3120240843@bit.edu.cn; yang\_yi@bit.edu.cn; yueyufeng@bit.edu.cn)

Digital Object Identifier XXXX.
}
}}

\maketitle

\begin{abstract}

Dynamic scene reconstruction represents a fundamental yet demanding challenge in computer vision and robotics.
{While recent progress in 3DGS-based methods has advanced dynamic scene modeling, obtaining high-fidelity rendering and accurate tracking in scenarios with substantial, intricate motions remains significantly challenging.}
To address these challenges, we propose PaMoSplat, a novel dynamic Gaussian splatting framework incorporating part awareness and motion priors.
Our approach is grounded in two key observations: 1) Parts serve as primitives for scene deformation, and 2) Motion cues from optical flow can effectively guide part motion.
{Specifically, PaMoSplat initializes by lifting multi-view segmentation masks into 3D space via graph clustering, establishing coherent Gaussian parts. For subsequent timestamps, we leverage a differential evolutionary algorithm to estimate the rigid motion of these parts using multi-view optical flow cues, providing a robust warm-start for further optimization.}
Additionally, PaMoSplat introduces an adaptive iteration count mechanism, internal learnable rigidity, and flow-supervised rendering loss to accelerate and optimize the training process.
\textcolor{black}{Comprehensive evaluations across diverse scenes, including real-world environments, demonstrate that PaMoSplat delivers superior rendering quality, improved tracking precision, and faster convergence compared to existing methods.} Furthermore, it enables multiple part-level downstream applications, such as 4D scene editing.

\end{abstract}


\section{Introduction}
\IEEEPARstart{3}{D} dynamic scene reconstruction plays a critical role in numerous computer vision and robotics applications, particularly in augmented and virtual reality (AR/VR) and real-to-sim transfer.
{More notably, the recent surge in tracking-any-point \cite{wen2023anypoint} or tracking-keypoint \cite{Rekep} policies highlights the pivotal role of dense tracking in advancing robot manipulation skills.}
\textcolor{black}{The core challenge in dynamic reconstruction lies in accurately modeling both the geometry and appearance of 3D scenes while maintaining persistent tracking of moving elements, enabling high-fidelity novel-view synthesis and temporally consistent motion analysis.}

The recent success of Neural Radiance Fields (NeRF) \cite{nerf} in Novel View Synthesis (NVS) has garnered significant attention, which leverages MLPs to represent static scenes.
Researchers have further extended the NeRF framework to model dynamics \cite{K-planes, hexplane, Dynibar, dynerf, tensor4d, D-nerf, HyperNeRF, Nerfplayer}.
While some efforts \cite{K-planes, hexplane, tensor4d, Hyperreel, qiao2023dynamic, lin2023high, mixvoxel} focus on accelerating vanilla NeRF, the computational burden of volumetric rendering and implicit representation continues to pose a significant challenge for the practical deployment of NeRF-like methods.

More recently, 3D Gaussian Splatting (3DGS) \cite{3DGS} has achieved real-time rendering of arbitrary views through differentiable point-based splatting with 3D Gaussians as rendering primitives. To extend 3DGS to dynamic scenarios, various methods have been introduced, including temporal extensions \cite{4k4d, 4d-gaussian-splatting, spacetime, 4d-rotor, yan20244d}, deformation fields \cite{sc-gs, Gaussian-flow, 4dgs, Deformable-Gaussians, lu2024geometry, superpoint-gs, GSprediction}, 
\textcolor{black}{and per-timestamp training \cite{d3dg, abou2024physically}.
Despite these advancements, current methods often fail to account for two fundamental physical principles, resulting in significant performance degradation when handling scenes with large, irregular motion patterns:}

\textbf{I) Parts serve as primitives for scene deformation.}
{In the physical world, parts naturally serve as coherent units with strong internal consistency, while connections between parts accommodate relative motion.
However, most existing 3DGS-based approaches \cite{4d-gaussian-splatting, 4dgs, spacetime, 4d-rotor, d3dg, Deformable-Gaussians} treat Gaussians as independent primitives, applying point-wise warping without structural constraints.
Although some methods  \cite{GSprediction, Gaussian-flow, lu2024geometry, superpoint-gs, sc-gs} enforce local spatial coherence, they struggle to model discontinuous motion at part boundaries, where the assumption of local rigidity is invalid.} Additionally, the absence of internal rigidity constraints leads to temporal inconsistencies and disorganized Gaussian behavior, especially in scenes with complex motions \cite{Panoptic_Studio}.

\textbf{II) Motion cues from optical flow can effectively guide part motion.}
\textcolor{black}{2D RGB sequences contain both appearance information for supervising Gaussian properties and valuable motion cues \cite{guo2024motion} that can inform part motion priors.} However, many algorithms \cite{4d-gaussian-splatting, 4dgs, spacetime, Gaussian-flow, superpoint-gs} tend to oversimplify the accuracy of Gaussian motion, often prioritizing the high fidelity of rendered images within the training views.
Although some contemporary works \cite{Gaussianflow, guo2024motion} integrate optical flow, the simple pixel-level supervised approach disrupts the optimization direction when the flow estimation is inaccurate.

To overcome these limitations, we propose \textbf{PaMoSplat}, a novel \textbf{Pa}rt-aware \textbf{Mo}tion-guided Gaussian \textbf{Splat}ting framework, which reconstructs iteratively dynamic Gaussian fields from multi-view videos.
The key technical challenge lies in both explicitly defining 3D parts from multi-view image sequences and effectively lifting 2D motion cues into 3D space. To tackle this, at the initial timestamp, PaMoSplat employs an off-the-shelf image mask predictor, SAM \cite{sam}, to obtain 2D segmentation masks, which are then associated with Gaussians from a standard 3DGS pipeline. 
Gaussian parts are further derived using a weighted graph clustering technique, which offers robustness against segmentation inaccuracies.
{For subsequent timestamps, PaMoSplat computes the 2D optical flow \cite{Raft} across all views and utilizes a differential evolutionary algorithm to infer the 3D prior motion of these Gaussian parts.} 
Based on the quality of the warm-start state, PaMoSplat adaptively determines the number of iterations for each subsequent timestamp.
Furthermore, PaMoSplat incorporates learnable internal rigidity for the parts and flow-supervised rendering loss to enhance the quality of dynamic modeling.
In summary, our contributions are as follows:
\begin{itemize}
    \item A novel part-aware motion-guided \textbf{dynamic Gaussian splatting framework PaMoSplat} is presented that facilitates dynamic novel view synthesis and dense tracking.
    \item \textbf{Gaussian parts with internal rigidity} are introduced as primitives for the deformation of the scene, ensuring higher temporal continuity throughout the process.
    \item {A differential evolutionary method based on multi-view optical flow is developed for \textbf{the prior motion of parts}, providing superior warm-start states that accelerate and optimize training.}
    \item PaMoSplat demonstrates \textbf{superior rendering and tracking accuracy} across a variety of datasets with complex, irregular motions, while its unique part-awareness facilitates downstream applications, such as 4D scene editing.
\end{itemize}


\section{Related Work}

\subsection{Static Novel View Synthesis}
\textcolor{black}{Recent years have witnessed remarkable progress in Neural Radiance Fields (NeRF) \cite{nerf}, with numerous extensions addressing increasingly complex scenarios. These include handling reflective surfaces \cite{cui2024aleth}, unbounded environments \cite{zhenxing2022switch}, and pose-free \cite{wang2023f2} or sparse-view settings \cite{guo2024depth}.}
Additionally, NeRF has been integrated with higher-dimensional understanding, such as fixed-set semantics \cite{wang2023semantic}, zero-shot learning \cite{OpenObj}, and surface reconstruction \cite{MACIM, LGSDF}.
However, most of these methods build on the vanilla NeRF, which remains computationally expensive.

\textcolor{black}{Parallel efforts have focused on accelerating NeRF-like rendering through two primary directions. The first stream optimizes NeRF's core architecture via light field representations \cite{gupta2024lightspeed} or improved sampling strategies \cite{ma2024hashpoint}. The second direction reduces computational complexity through spatial partitioning techniques, including multi-subfield approaches \cite{you2024generative} and hybrid explicit-implicit representations \cite{inst-ngp}. The recent introduction of 3DGS \cite{3DGS} represents a significant breakthrough, achieving real-time novel view synthesis through differentiable point-based rendering with 3D Gaussians.} However, all aforementioned methods remain fundamentally limited to static scene representations.

\begin{figure*}[!t]\centering
	\includegraphics[width=16cm]{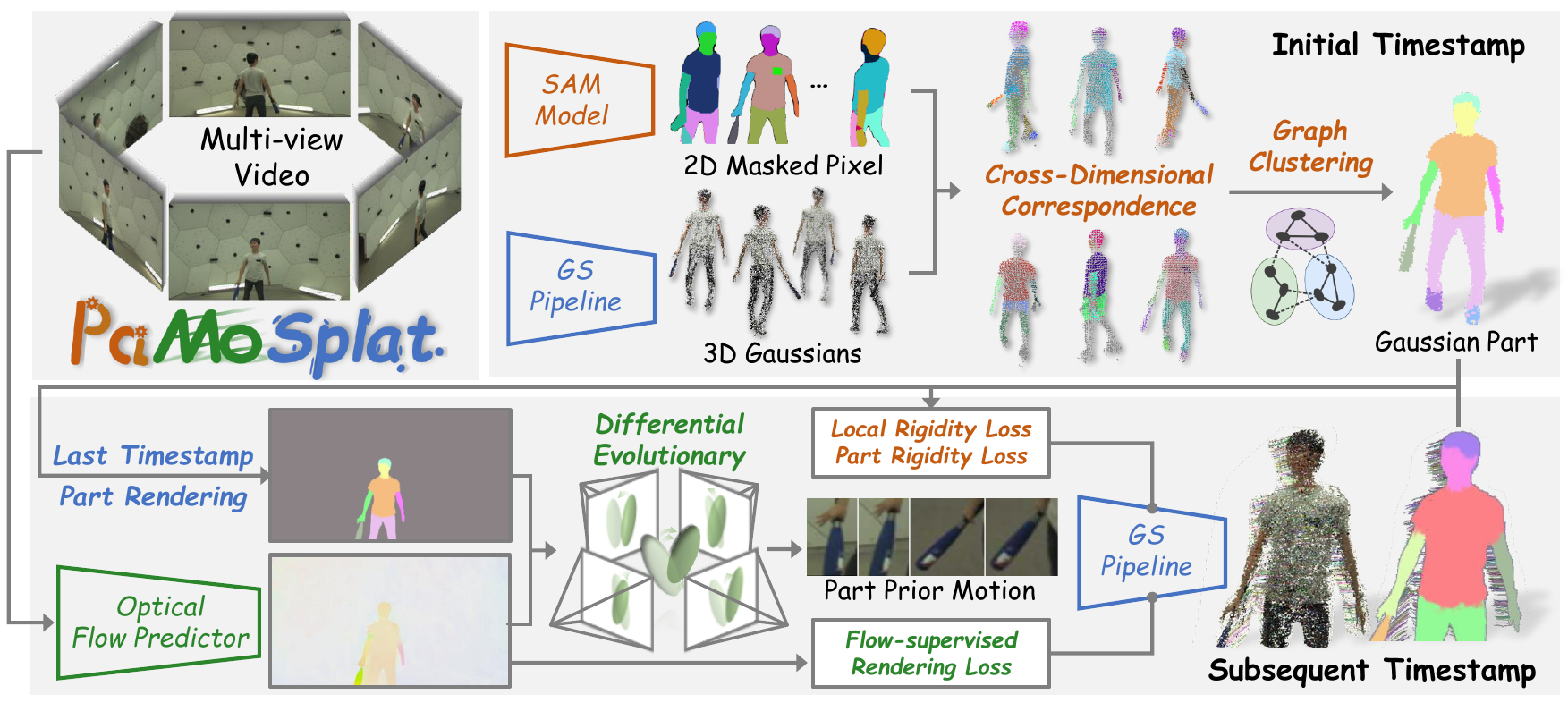}
	\caption{ \textbf{Overview of the PaMoSplat pipeline.} PaMoSplat introduces Gaussian parts in the initial timestamp, which are shifted in subsequent timestamps based on optical flow. Additionally, PaMoSplat incorporates some rigid constraints and flow-supervised losses to further refine the dynamic scene representation.} 
	\label{framework} 
\end{figure*}

\subsection{Dynamic Neural Radiance Field}

A straightforward approach to extend NeRF to dynamic scenarios is by introducing a time dimension \cite{K-planes, hexplane, gao2021dynamic, xian2021space, Dynibar, dynerf, mixvoxel, lin2023high, tensor4d, ngnerf}. 
Among these, DyNeRF \cite{dynerf} introduces a novel time-conditional neural radiation field that leverages a compact set of latent codes to capture scene dynamics. \cite{gao2021dynamic} jointly trains a static NeRF and a dynamic NeRF. 
\textcolor{black}{Video-NeRF \cite{xian2021space} incorporates estimated scene depth to constrain the time-varying geometry.}
K-Planes \cite{K-planes}, HexPlanes \cite{hexplane} and Tensor4D \cite{tensor4d} factorize 4D spatio-temporal features into 2D planes.

Another popular approach is to construct a canonical field along with a deformation field \cite{D-nerf, guo2022neural, liu2023robust, Nerfies, HyperNeRF, Nerfplayer, Dynamicsurf, lin2025dynamic} or a flow field \cite{somraj2024factorized, guo2023forward, Evdnerf, Dynpoint} to capture scene dynamics. Among these, Nerfies \cite{Nerfies} optimizes an additional continuous volume deformation field to warp each observation into a standard 5D NeRF. Similarly, D-NeRF \cite{D-nerf} divides the learning process into two stages: encoding the scene into a canonical space and mapping it to time-specific deformations. 
\textcolor{black}{To address the challenges in modeling topological changes, HyperNeRF \cite{HyperNeRF} lifts the NeRF to a higher dimensional hyperspace.} 

Although some above methods employ feature planes \cite{K-planes, hexplane, tensor4d}, neural voxels \cite{guo2022neural, Dynamicsurf, mixvoxel, lin2023high}, model decomposition \cite{Nerfplayer, qiao2023dynamic}, or efficient sampling strategies \cite{Hyperreel, dynerf} to improve training and inference efficiency, NeRF-based methods still face challenges in achieving real-time rendering, and their inherent implicit representations limit interpretability and adaptability.

\subsection{Dynamic 3D Gaussian Splatting}

{Following the breakthrough of 3DGS \cite{3DGS}, dynamic extensions broadly fall into two modeling paradigms: Continuous Spatiotemporal Fitting and Per-timestamp Training.}

{\textbf{Continuous Spatiotemporal Fitting.} This category approximates scenes as continuous functions within bounded volumes. Some methods construct temporal GS: 4K4D \cite{4k4d} builds 4D point clouds for fast rasterization, STGaussian \cite{spacetime} adds temporal opacity, while 4D-Gaussian \cite{4d-gaussian-splatting} and 4DRotorGS \cite{4d-rotor} optimize anisotropic XYZT Gaussians. Others learn continuous deformation fields to warp canonical Gaussians using varied backbones: Deformable-GS \cite{Deformable-Gaussians} uses coordinate-based MLPs, while E-D3DGS \cite{ed3dgs} employs per-Gaussian embeddings for finer control. Gaussian-Flow \cite{Gaussian-flow} utilizes a dual-domain model, and 4DGS \cite{4dgs} and SC-GS \cite{sc-gs} adopt HexPlane-like \cite{hexplane} voxels or control points. To improve fidelity, ST-4DGS \cite{st4dgs} adds motion-aware shape regularization, and Swift4D \cite{swift4d} utilizes 4D hash encoding in a divide-and-conquer framework. 
Despite technical differences, these methods reduce dynamic reconstruction to an appearance-fitting problem via photometric loss. The implicit assumption of continuity often struggles with large structural changes or distinct object boundaries and tends to overfit time-varying characteristics within specific spatiotemporal regions.}

{\textbf{Per-timestamp Training.} To improve extensibility, the second category adopts per-timestamp training. D-3DGS \cite{d3dg} pioneered incremental updates with physical priors, while 3DGStream \cite{3dgstream} utilizes a two-stage pipeline with neural transformation caching. HiCoM \cite{hicom} further accelerates this via a hierarchical coherent motion mechanism exploiting local consistency. The primary limitation of these methods lies in treating Gaussians as independent primitives, which deviates from physical motion principles. Physical motion inherently occurs as coordinated rigid or semi-rigid bodies. Without structural awareness, optimization is prone to local minima, leading to tracking drift and temporal jitter.}

{In contrast, PaMoSplat is motivated by the insight that scenes are composed of physically driven, coordinated parts. While utilizing per-timestamp training, PaMoSplat models Gaussian parts as coherent motion units and introduces flow-guided prior motion. It is built upon two visual networks: SAM and RAFT. SAM \cite{sam}, constructed with a heavy Vision Transformer (ViT) encoder, demonstrates exceptional zero-shot generalization capabilities. This enables PaMoSplat to explicitly group discrete Gaussians into semantically coherent Gaussian parts across arbitrary scenes. RAFT \cite{Raft} constructs 4D multi-scale correlation volumes for all pixel pairs and iteratively updates the flow field via recurrent GRU-based units. Within our framework, RAFT provides explicit motion cues to warm-start Gaussian parts, effectively mitigating local minima. Notably, these backbone networks can be replaced with functionally equivalent alternatives, and experiments demonstrate that PaMoSplat remains highly robust to such substitutions.}

\section{Method}

\subsection{PaMoSplat Framework} \label{frame}

Given a set of multi-view videos $\{\mathcal{I}_{t,v}\}$ (where $\mathcal{I}_{t,v}$ is the color image captured at timestamp $t$ from camera view $v$ with intrinsic parameters ${K_{v}}$ and extrinsic parameters ${E_{v}}$), PaMoSplat extends the vanilla static 3D Gaussian Splatting pipeline to handle dynamic 3D scenes in a temporally consistent manner. 
{This formulation enables PaMoSplat to iteratively refine the Gaussian field $\mathcal{S}_{t}$ over time, which is composed of 3D Gaussian ellipsoids parameterized by:}
\begin{enumerate}
    \item \textit{3D center:} \quad ${{\mu}_{t}} = ({{x}_{t}},{{y}_{t}},{{z}_{t}})\in {{\mathbb{R}}^{3}}$
    \item \textit{3D rotation:} \quad ${{{{q}}}_{t}} = (q{{w}_{t}},q{{x}_{t}},q{{y}_{t}},q{{z}_{t}})\in {{\mathbb{R}}^{4}}$
    \item \textit{RGB color:} \quad ${c}_t = \text{(}r_t,g_t,b_t\text{)}\in {{\mathbb{R}}^{3}}$
    \item \textit{3D size (scaling factor):} \quad $\text{s} = \text{(}sx,sy,sz\text{)}\in {{\mathbb{R}}^{3}}$
    \item \textit{Opacity logit:} \quad $o\in [0,1]$
    \item \textit{Part ID:} \quad $pi\in \mathbb{Z}$
\end{enumerate}

{Among these parameters, only the center position $\mu_t$, rotation quaternion ${{{{q}}}_{t}}$, and color ${c}_t$ (optional to simulate lighting variations) evolve over time, while all other parameters remain static after initialization.} Unlike standard 3D Gaussian Splatting, we introduce an additional parameter: the part ID $pi$, which associates each Gaussian with a specific part. Note that the concept of a part here is broad, encompassing even small objects as complete parts.

As illustrated in Fig. \ref{framework}, PaMoSplat comprises two primary components. At the initial timestamp, we cluster the Gaussians into distinct Gaussian parts. Then, at subsequent timestamps, we estimate the prior motion of these Gaussian parts and introduce rigidity constraints along with a flow-supervised rendering loss. These processes yield a progressively evolving Gaussian field, $\mathcal{S}_{t}$.

\begin{figure}[!t]\centering
	\includegraphics[width=8cm]{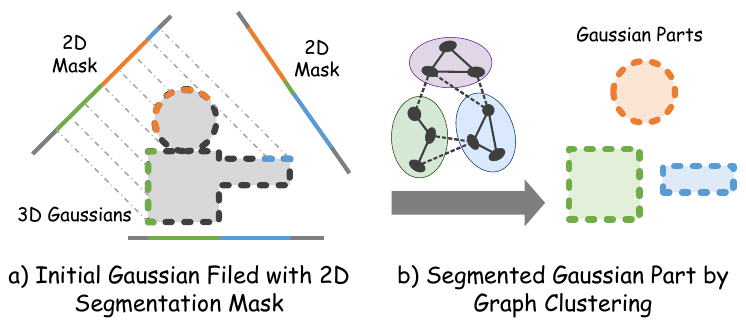}
	\caption{ \textbf{Process of Gaussian part generation.} The initialized 3D Gaussian field $\mathcal{S}_{0}$ and the multi-view 2D segmentation masks $\{m_{v}^{i}\}$ serve as inputs, allowing for the assignment of part IDs $pi$ to Gaussians through cross-dimensional correspondence and graph clustering.} 
	\label{part_gen} 
\end{figure}

\subsection{Initial Timestamp} \label{initial}



At the initial timestamp $t=0$, we process the multi-view images  $\{\mathcal{I}_{v}\}$ (abbreviation for $\{\mathcal{I}_{0,v}\}$) through two parallel pipelines. The first pipeline follows the standard 3DGS approach to generate the initial Gaussian field $\mathcal{S}_{0}$, where we initialize from a sparse point cloud and optimize for 10,000 iterations with Gaussian densification.
{Concurrently, the second pipeline employs SAM \cite{sam} to produce per-view segmentation masks $\{m_{v}^{i}\}$.} We employ SAM's automatic segmentation mode with a post-processing strategy that layers smaller masks atop larger ones to prevent overlapping regions.
Through subsequent correspondence and clustering operations (visualized in Fig. \ref{part_gen}), these 2D masks are lifted to 3D space to form coherent Gaussian parts.

\vspace{1em}
\noindent
\textbf{Pixel-Gaussian Cross-Dimensional Correspondence.}
To efficiently align the 2D masks ${m_{v}^{i}}$ with the 3D Gaussian field ${\mathcal{S}_{0}}\triangleq \{{{g}^{j}}\}$, we introduce a depth-guided Cross-Dimensional Correspondence approach $\operatorname{Cor}(\cdot ,\cdot )$. This method leverages NeRF-like pixel rays to identify the nearest Gaussian along each ray's path, searching near the rendering depth. 
\textcolor{black}{When the depth difference is less than a defined threshold ${\theta}_{depth}$ (scaled proportionally to the scene size, set to 0.1m for real-world-scale datasets in our experiments), the Gaussian ${g}^{j}$ with center ${\mu}^{j}$ is deemed visible to $v$th camera, forming the set $\mathcal{V}_v$:} 
\textcolor{black}{
\begin{equation}
\begin{aligned}
    \label{cdc_gs}
    \mathcal{V}_v &= \operatorname{Cor}({S}_{0},{\mathcal{D}}_{v})\\
    &= \left\{{g}^{j}\left| | ({{E}_{v}}{{{{\mu }}}^j})_z-{{{{\mathcal{D}}}}_{v}}[{{K}_{v}}{{E}_{v}}{{{{\mu }}}^j}] |<{{\theta }_{depth}} \right. \right\}
\end{aligned}
\end{equation}} 
\textcolor{black}{where ${\mathcal{D}}_{v}$ represents the rendered depth image of the Gaussian field $\mathcal{S}_0$ for $v$th camera, with intrinsic parameters $K_v$ and extrinsic parameters $E_v$. 
Here, $({{E}_{v}}{{{{\mu }}}^j})_z$ denotes the depth of the Gaussian center $\mu^j$ in the $v$th camera's coordinate frame, and ${{K}_{v}}{{E}_{v}}{{{{\mu }}}^j}$ is its corresponding pixel projection.}

\vspace{1em}
\noindent
\textbf{Graph Clustering.}
The 2D masks ${m_{v}^{i}}$ provide part segmentation of the visible Gaussians. To generate Gaussian parts in 3D, these masks must be aggregated across all views. This is accomplished by initializing a fully connected graph $\mathcal{G}$, where the nodes correspond to Gaussians ${g^j}$ and the edges represent pairwise connections. 
{The edge weights $\omega({g}^{j},{g}^{k})$ are defined as the ratio between the number of times the two connected Gaussians ${g}^j$ and ${g}^k$ appear together in the same masks $m_{v}^{i}$ and the total number of their co-visible views $v$:}
\begin{equation}
\begin{aligned}
    \label{graph}
    \mathcal{G}=\left( {{{{g}}}},\omega ( {g}^{j},{g}^{k} ) \right)
\end{aligned}
\end{equation}
\textcolor{black}{
\begin{equation}
\begin{aligned}
    \label{graph_weight}
    \omega \left( {{g}^{j}},{{g}^{k}} \right)=\frac{\sum\limits_{v}{\sum\limits_{i}{\mathbf{1}_{\left( {{g}^{j}}\in {\mathcal{V}_v[{m_{v}^{i}}]} \right)}\cdot \mathbf{1}_{\left( {{g}^{k}}\in {\mathcal{V}_v[{m_{v}^{i}}]} \right)}}}}{\sum\limits_{v}{\mathbf{1}_{\left( {{g}^{j}}\in {\mathcal{V}_v} \right)}\cdot \mathbf{1}_{\left( {{g}^{k}}\in {\mathcal{V}_v} \right)}}}
\end{aligned}
\end{equation}}
\textcolor{black}{where $\mathbf{1}_{(\cdot)}$ denotes the indicator function, ${\mathcal{V}_v[{m_{v}^{i}}]}$ represents  the visible Gaussian primitives within the mask ${m_{v}^{i}}$, and $\mathcal{V}_v$ denotes all visible Gaussians in the $v$th camera, as defined in Eq. \eqref{cdc_gs}.}
We then apply the Louvain algorithm \cite{Louvain} for graph clustering on the nodes of $\mathcal{G}$, with each cluster corresponding to a distinct Gaussian part $\mathcal{P}^{pi}$ and assigning part ID $pi$. These IDs are fixed in the subsequent timestamps to explicitly define the part associations.

\subsection{Subsequent Timestamp} \label{subsequent}


\textcolor{black}{In PaMoSplat, the Gaussian field $\mathcal{S}_{t}$ at each subsequent timestamp $t>0$ is optimized through an iterative refinement based on the last one $\mathcal{S}_{t-1}$, updating the Gaussian centers $\mu_t$, rotation quaternions ${{{{q}}}_{t}}$, and colors ${c}_t$. 
A distinctive aspect of our approach is the integration of optical flow to enable efficient preliminary deformation of the Gaussian field. This is achieved by warping each Gaussian part as an independent rigid unit, establishing an effective initialization for the formal backpropagation-based 3DGS optimization.}

\vspace{1em}
\noindent
\textbf{Last Timestamp Part Rendering.} 
{To establish a direct correspondence with 2D optical flow, we render the Gaussian parts $\mathcal{P}^{pi}$ into 2D part images for each viewpoint.} In practice, each part is randomly assigned a distinct color, following the same rendering scheme as RGB color $c$. The resulting image represents the part segmentation in the view $v$ for the Gaussian field $\mathcal{S}_{t-1}$ at the last timestamp, which is defined as $\{m^{pi}_{v,t-1}\}$.

\vspace{1em}
\noindent
\textbf{Optical Flow Prediction.} Image sequences provide rich motion cues, especially through optical flow representations. 
{We utilize RAFT \cite{Raft} to compute both forward $\mathcal{O}_{v,t}$ and backward ${\mathcal{O}}^{-}_{v,t}$ flow fields.}
The forward flow $\mathcal{O}_{v,t}$ directly guides the deformation of the last Gaussian field $\mathcal{S}_{t-1}$ to the current timestamp $t$, while both $\mathcal{O}_{v,t}$ and ${\mathcal{O}}^{-}_{v,t}$ jointly serve as supervision for the rendering loss of attention.

\begin{figure}[!t]\centering
	\includegraphics[width=8cm]{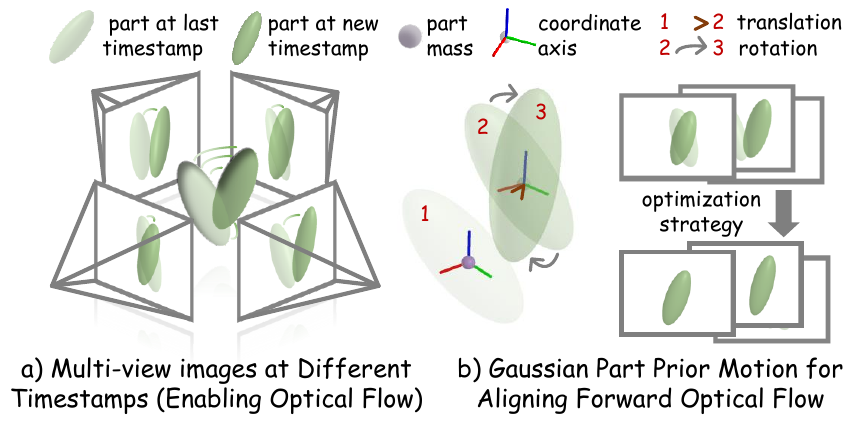}
	\caption{ \textbf{Process of Gaussian part prior motion based flow.} For each part $\mathcal{P}^{pi}$, we employ a differential evolutionary algorithm to compute the optimal translation $\Delta^{pi}$ and rotation $\Omega^{pi}$, ensuring that the computed optical flow $\hat{\mathcal{O}}_{v,t}$ aligns with the ground truth flow $\mathcal{O}_{v,t}$. \textbf{Note}: This figure utilizes a combination of last and new timestamp observations as a surrogate for optical flow. \textcolor{black}{The order of translation and rotation is irrelevant, provided that the rotation is applied around the current center of mass.}} 
	\label{DE} 
\end{figure}

\vspace{1em}
\noindent
\textbf{Part Prior Motion.} \textcolor{black}{While single-view optical flow alone cannot determine 3D motion, the integration of multi-view flow observations enables robust 3D motion estimation. We cast this as an optimization problem where we estimate six degrees of freedom (6-DoF) for each Gaussian part $\mathcal{P}^{pi}$: three translational components $\Delta^{pi} = (\Delta^{pi}_x,\Delta^{pi}_y,\Delta^{pi}_z$) and three rotational components $\Omega^{pi} = (\Omega^{pi}_x,\Omega^{pi}_y,\Omega^{pi}_z$) around the center of mass $\mathcal{C}^{pi}_{t-1}$ of Gaussian part $\mathcal{P}^{pi}$:}
\textcolor{black}{
\begin{equation}
\begin{aligned}
  \label{mass}
  \mathcal{C}_{t-1}^{pi}=\left\{ \left. \operatorname{mean}(\mu _{t-1}^{j}) \right|{{g}^{j}}\in {\mathcal{P}^{pi}} \right\} 
\end{aligned}
\end{equation}}
\textcolor{black}{The optimization objective is formulated to minimize the discrepancy between the rendered optical flow $\hat{\mathcal{O}}_{v,t}$ and the observed ground truth flow $\mathcal{O}_{v,t}$.} For part $\mathcal{P}^{pi}$, the problem can be formalized as:
\begin{equation}
\begin{aligned}
  \label{single_loss}
  \underset{{{ \Delta^{pi},\Omega^{pi}}}}{\mathop{\text{minimize}}}\quad \sum\limits_{v} \left\| \hat{\mathcal{O}}_{v,t}[m^{pi}_{v,t-1}] - \mathcal{O}_{v,t}[m^{pi}_{v,t-1}] \right\| 
\end{aligned}
\end{equation}
In this formulation, $\hat{\mathcal{O}}_{v,t}$ is computed based on the spatial transformation projected onto the image:
\begin{subequations}
\begin{align}
  \label{cal_flow}
  {\hat{\mathcal{O}}_{v,t}}[m_{v,t-1}^{pi}]=\left\{ {{K}_{v}}{{E}_{v}}(\hat{\mu }_{t}^{j}-\mu _{t-1}^{j}) \right\} \\ 
  \label{transform_gaissoan}
  \textcolor{black}{
 \hat{\mu }_{t}^{j}={{R}^{pi}}(\mu _{t-1}^{j}-\mathcal{C}_{t-1}^{pi})+\mathcal{C}_{t-1}^{pi}+{{\Delta }^{pi}} 
 }
\end{align}
\end{subequations}
where ${R}^{pi}$ is the rotation matrix derived from $\Omega^{pi}$. \textcolor{black}{Fig. \ref{DE}  provides a visual illustration of this process.} 
To solve Eq. \eqref{single_loss}, we employ a multi-parameter Differential Evolution (DE) algorithm to efficiently obtain a globally optimal solution.
{Specifically, we use Eq. \eqref{single_loss} as the objective function, apply the `best/1/bin' search strategy, constrain the search ranges for translation $\Delta^{pi}$ and rotation $\Omega^{pi}$ to $\pm0.2$m and $\pm20^\circ$, respectively, set the population size to 2, and limit the maximum number of iterations to 25.} The final iteration results are then applied to the Gaussian part $\mathcal{P}^{pi}$ to realize the prior motion.

\begin{figure}[!t]\centering
	\includegraphics[width=7cm]{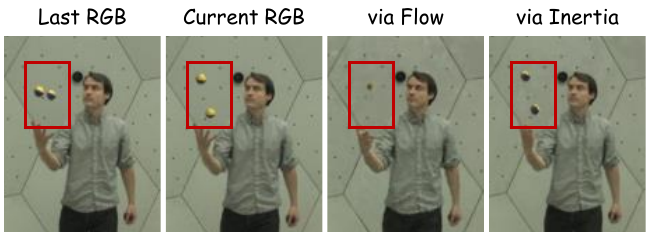}
	\caption{ \textbf{An example of failed prior motion estimation using optical flow} demonstrates balls being displaced to unintended locations. In this case, the proposed \textbf{part inertia mechanism} helps guide these parts to the correct position.} 
	\label{inertia} 
\end{figure}

Moreover, to tackle potential failures of optical flow, we introduce an inertia mechanism. In contrast to \cite{d3dg}, where the inertia is applied independently to each Gaussian, we apply it to the entire part.
We propose a part inertia mechanism that models the motion of each part as a whole relative to its previous state. We decompose inertia into translational inertia $\Theta ^{pi}_{tran}$ and rotational inertia $\Theta ^{pi}_{rota}$.
This calculation utilizes the centers of each Gaussian within part $\mathcal{P}^{pi}$ at the last and penultimate timestamps, denoted as $\{\mu^{j}_{t-1}\}$ and $\{\mu^{j}_{t-2}\}$, respectively, \textcolor{black}{ along with the center of mass $\mathcal{C}^{pi}_{t-1}$ and $\mathcal{C}^{pi}_{t-2}$ as defined in Eq. \eqref{mass}.} The translational inertia $\Theta ^{pi}_{tran}$ can be derived directly from the offset of the center of mass:
\begin{equation}
\begin{aligned}
  \label{tran_iner}
  \Theta ^{pi}_{tran} = \mathcal{C}^{pi}_{t-1}- \mathcal{C}^{pi}_{t-2}
\end{aligned}
\end{equation}
{For the rotational inertia component, $\Theta ^{pi}_{rota}$, we model the rotation around the part's center of mass $\mathcal{C}^{pi}_{t-1}$ and $\mathcal{C}^{pi}_{t-2}$.} To obtain this, we calculate the covariance matrix of each Gaussian’s offset relative to the center of mass, followed by performing a singular value decomposition $\operatorname{SVD}(\cdot)$:
\begin{subequations}
\begin{align}
    \label{offset_t-1}
  {{O}_{t-1}} &=\{\mu _{t-1}^{j}-\mathcal{C}_{t-1}^{pi}|{{g}^{j}}\in {\mathcal{P}^{pi}}\} \\ 
  \label{offset_t-2}
 {{O}_{t-2}} &=\{\mu _{t-2}^{j}-\mathcal{C}_{t-2}^{pi}|{{g}^{j}}\in {\mathcal{P}^{pi}}\}
    \\
  \label{svd}
 \text{U }\!\!\Sigma\!\!\text{ }{{\text{V}}^{\text{T}}} &=\operatorname{SVD}(O_{t-2}^{\text{T}}{{O}_{t-1}}) \\
  \label{rata}
 \Theta ^{pi}_{rota} &= \text{U}{{\text{V}}^{\text{T}}}
\end{align}
\end{subequations}

where $\Theta^{pi}_{rota}$ is the rotation matrix.

We substitute the computed $\Theta^{pi}_{tran}$ and $\Theta^{pi}_{rota}$ as replacements for the parameters $\Delta^{pi}$ and $\Omega^{pi}$ in the prior motion estimation of the Gaussian part $\mathcal{P}^{pi}$, as defined in Eq. \eqref{single_loss}. 
\textcolor{black}{Optical flow failures are detected by evaluating RGB differences within the part rendering mask $\{m^{pi}_{v,t}\}$: significant discrepancies indicate that parts guided by optical flow have not reached their expected positions.
Fig. \ref{inertia} illustrates an example of motion supported by inertia.}


\vspace{1em}
\noindent
\textbf{Adaptive iteration count mechanism.}
Through the part prior motion mechanism, we have obtained more accurate initial states for most timestamps. \textcolor{black}{To reduce unnecessary computational overhead on frames that have already been well-optimized through prior motion or exhibit minimal motion, we introduce a simple yet effective adaptive iteration strategy.} Specifically, we render the Gaussian field after prior motion to each camera view $v$ and compare the rendered images with the real images. \textcolor{black}{The mean RGB difference ${d}_{pixel}$ serves as a key indicator for determining the number of iterations $\Upsilon_t$:}
\begin{equation}
\begin{aligned}
  \label{adaptive_iteration}
  {{\Upsilon  }_{t}}=\operatorname{clip}(\varepsilon \cdot {{d}_{pixel}},{{\Upsilon}_{\min}},{{\Upsilon  }_{\max }})
\end{aligned}
\end{equation}
\textcolor{black}{where $\varepsilon$ is the multiplication factor and $\operatorname{clip}(\cdot)$ is a clamping function. In the experiments, we set $\varepsilon$ to $10^5$, ${\Upsilon}_{\min}$ to 1,500, and ${\Upsilon}_{\max}$ to 2,000.}

\begin{figure}[!t]\centering
	\includegraphics[width=7cm]{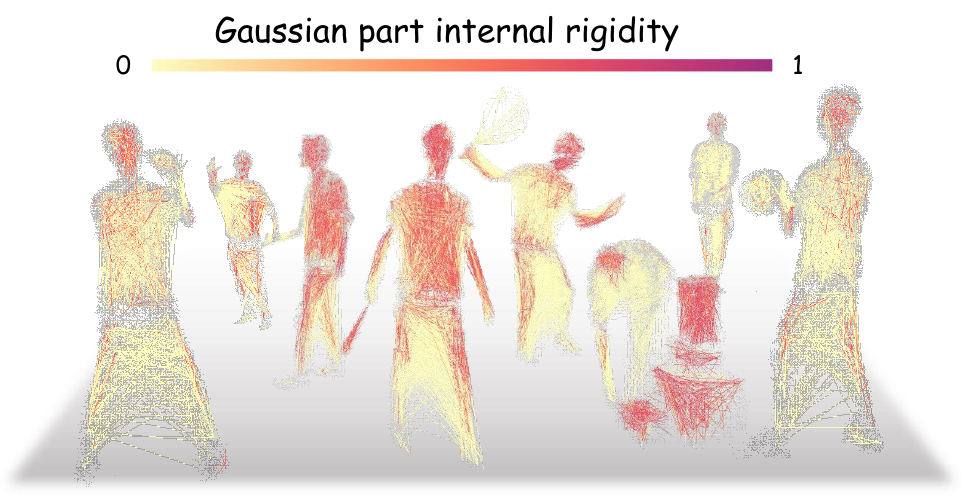}
	\caption{ \textbf{Learnable internal rigidity of Gaussian part.} Although the legs are treated as a whole part, PaMoSplat can discern that the internal rigidity within one leg is greater than that between the two legs. The lines in the image connect Gaussians to their anchors $\mathcal{A}(g^{j})$, with color indicating the rigidity $\mathcal{W}_t \in [0,1]$.} 
	\label{part_rigid} 
\end{figure}

\vspace{1em}
\noindent
\textbf{Flow-supervised Rendering Loss.}
After completing the prior motion estimation for each part, the standard Gaussian splatting optimization process is initiated.
Similar to most 3DGS-based methods, we incorporate L1 color loss $\mathcal{L}_{1}$ and D-SSIM terms $\mathcal{L}_{D}$. Additionally, we introduce a flow-supervised loss $\mathcal{L}_{O}$ to emphasize highly dynamic regions:
\begin{subequations}
\begin{align}
  \label{color_loss}
  {\mathcal{L}_{Img}} &={{\lambda }_{c}}{\mathcal{L}_{1}}+{{\lambda }_{s}}{\mathcal{L}_{D}}+{{\lambda }_{o}} \mathcal{L}_{O} \\
  \label{flow_weigt}
  \mathcal{L}_{O} & = \left| \left( {{{\hat{\mathcal{I}}}}_{v,t}}-{\mathcal{I}_{v,t}} \right) \cdot \operatorname{Norm}\left( \left\| {\mathcal{O}_{v,t}} \right\|+\left\| \mathcal{O}_{v,t}^{-} \right\| \right) \right|
\end{align}
\end{subequations}
where ${{{\hat{\mathcal{I}}}}_{v,t}}$ is the rendered image, ${{{{\mathcal{I}}}}_{v,t}}$ is the real image and $\operatorname{Norm}(\cdot)$ is the normalization function.

\vspace{1em}
\noindent
\textbf{Learnable Internal Rigidity Loss.} 
Some methods \cite{d3dg, Deformable-Gaussians} incorporate KNN rigidity loss $\mathcal{L}_{Loc-rigid}$ to enforce local consistency constraints.
{However, this assumption is invalid at part boundaries characterized by discontinuous relative motion (e.g., the interaction between a hand and a basketball).
To mitigate this issue, we introduce a learnable part-level internal rigidity loss $\mathcal{L}_{Part-rigid}$, designed to maintain structural coherence throughout extended motion sequences.}
Specifically, each Gaussian $g^{j}$ randomly selects a subset of other Gaussians within the same part as anchors $\mathcal{A}(g^{j})$, with $\mathcal{L}_{part-rigid}$ penalizing changes in their relative distances. Since parts are not fully rigid (e.g., large and small sections of an arm), we propose intra-part learnable rigidity weights.
The internal rigidity factor, $\mathcal{W}_t$, continuously tracks the change in the relative distance between $g^{j}$ and $\mathcal{A}(g^{j})$:
\begin{subequations}
\begin{align}
  \label{distance_t-1}
  {{D}_{t-1}}&=\left\{ \left\| \mu _{t-1}^{j}-\mu _{t-1}^{k} \right\||{{g}^{k}}\in \mathcal{A}({{g}^{j}}) \right\} \\
  \label{distance_t-2}
 {{D}_{t-2}}&=\left\{ \left\| \mu _{t-2}^{j}-\mu _{t-2}^{k} \right\||{{g}^{k}}\in \mathcal{A}({{g}^{j}}) \right\} \\
  \label{delta}
 \Delta D &= |{{D}_{t-1}}-{{D}_{t-2}}|\\
  \label{rigidity}
 {\mathcal{W}_{t}}&={\mathcal{W}_{t-1}}+\alpha \max (1-\frac{\Delta D}{\delta },0)-\beta \max (\frac{\Delta D}{\delta }-1,0) \\
  \label{clip}
 {\mathcal{W}_{t}}&=\operatorname{clip}({\mathcal{W}_{t-1}},0,1)
\end{align}
\end{subequations}
where $\delta$ is the distance stability threshold, $\alpha$ and $\beta$ are the rigidity growth and decay rates ($\alpha<\beta$). 
\textcolor{black}{In the experiments, we set $\delta$ to $10^{-3}$m, $\alpha$ to 0.02, and $\beta$ to 0.2.}
The solved ${\mathcal{W}_{t}}$ is used to guide the part rigidity loss $\mathcal{L}_{part-rigid}$:
\begin{equation}
\begin{aligned}
  \label{part_rigid_loss}
  \mathcal{L}_{part-rigid} = {\mathcal{W}_{t}}|D_{t}-D_{t-1}|
\end{aligned}
\end{equation}
Fig. \ref{part_rigid} shows some results of the visualization of the internal rigidity.

\section{Experiment}


\subsection{Experimental Settings} \label{ES}

\vspace{1em} \noindent  \textbf{Implementation Details.}  
Our implementation primarily utilizes the PyTorch framework \cite{Pytorch} and is tested on a single RTX 4090 GPU. We initiate the process by sparsifying the depth images to create the initial scene and perform densification \cite{3DGS} only at the initial timestamp. 
\textcolor{black}{In the PanopticSports dataset, PaMoSplat applies the same foreground segmentation loss as D-3DGS \cite{d3dg}. At the initial timestamp, to ensure each Gaussian consistently represents a small specific area of physical space,} we constrain its size to no more than 3cm in any direction. Although Gaussian colors are non-fixed at subsequent timestamps, we impose a color consistency loss between consecutive timestamps as a constraint, following \cite{d3dg}, under the assumption that scene lighting changes are generally gradual.
{Hyperparameters, including depth thresholds and search ranges, are kept consistent across all scenes and selected based on empirical experience.}

\vspace{1em} \noindent  \textbf{Baseline implementation.}
\textcolor{black}{We select a diverse set of baselines, including 2 NeRF-based methods (\underline{HyperReel} \cite{Hyperreel}, \underline{K-Planes} \cite{K-planes}), 3 temporal GS methods (\underline{4D-Gaussians} \cite{4d-gaussian-splatting}, \underline{STGaussian} \cite{spacetime}, \underline{4DRotorGS} \cite{4d-rotor}), 2 deformation-field GS methods (\underline{4DGS} \cite{4dgs}, \underline{Deformable-GS} \cite{Deformable-Gaussians}), and 3 per-timestamp training GS methods (\underline{3DGStream} \cite{3dgstream}, \underline{3DGS-O} \cite{3DGS}, \underline{D-3DGS} \cite{d3dg}). Notably, 3DGS-O is a variant of the original 3DGS \cite{3DGS} designed specifically for iterative dynamic modeling.
All experiments are conducted using the official implementations of each baseline.} Since methods like 4D-Gaussian \cite{4d-gaussian-splatting}, STGaussian \cite{spacetime}, 4DGS \cite{4dgs}, and Deformable-GS \cite{Deformable-Gaussians} are not extensively evaluated on large-motion datasets such as PanopticSports in their original papers, we preprocess all datasets to the required formats. The simple qualitative results (Fig. 17) in 4DGS paper \cite{4dgs} align with our experimental findings.

\begin{figure}[!t]\centering
	\includegraphics[width=8cm]{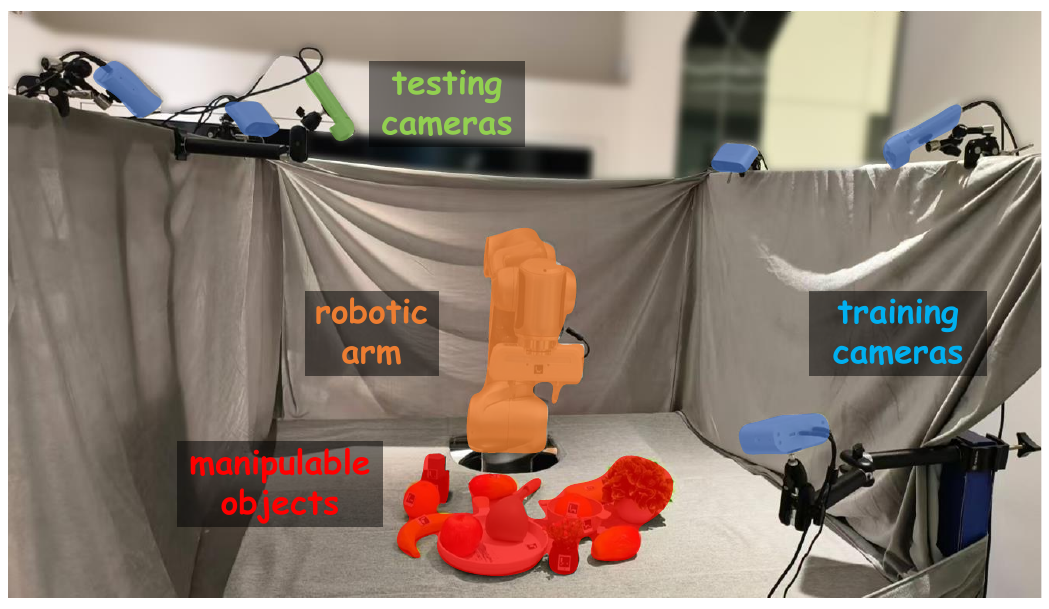}
	\caption{ \textbf{A multi-view video capture platform for self-captured dataset.} Five training cameras and one test camera, each with distinct views, simultaneously capture the process of object manipulation by the robot arm.} 
	\label{self_captured} 
\end{figure}

\begin{figure*}[!t]\centering
	\includegraphics[width=18.2cm]{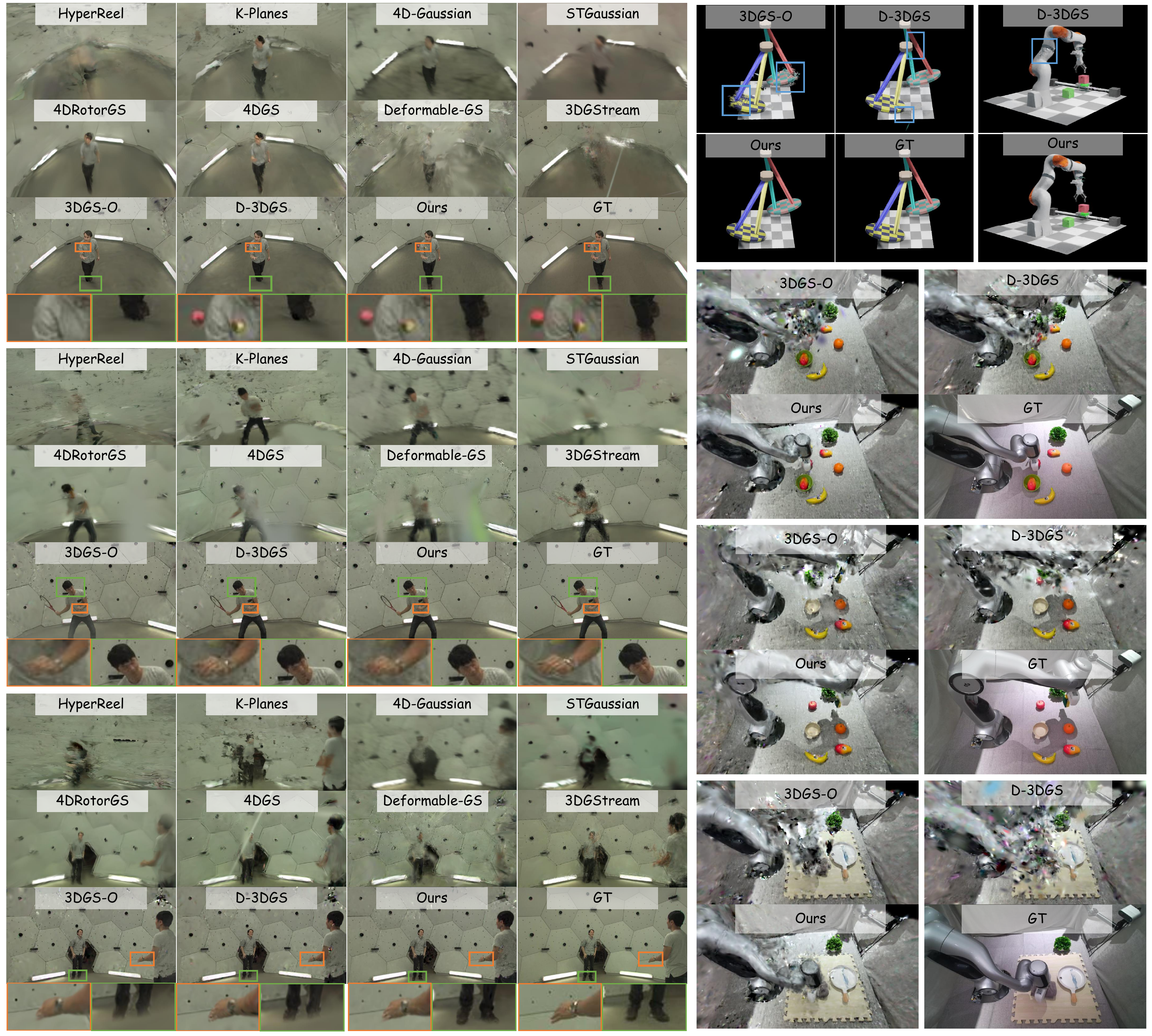}
	\caption{ \textcolor{black}{\textbf{Qualitative comparisons of novel view synthesis.} We zoom in on the main comparison baselines to highlight details. With prior motion and optical flow supervision, PaMoSplat demonstrates a significant advantage in modeling highly dynamic elements, even with the sparse view  (5 training cameras in the self-captured dataset displayed in the lower right corner).}}
	\label{render_main} 
\end{figure*}

\begin{table*}[!t]
\scriptsize
\centering
\caption{
\textcolor{black}{\textbf{Per-scene quantitative comparisons of novel view synthesis on PanopticSports  dataset \cite{d3dg}.}
The results are based on official code implementations of each corresponding method. Our results achieves the highest metrics (\textbf{PSNR $\uparrow$, SSIM $\uparrow$, LPIPS $\downarrow$}).}
}
\label{cmu_table}
\renewcommand\arraystretch{1.2}
\setlength{\tabcolsep}{0.365mm}
{\begin{tabular}{l|ccc|ccc|ccc|ccc|ccc|ccc|ccc}
\toprule
\multicolumn{1}{c|}{\multirow{2}{*}{\textbf{Methods}}} & \multicolumn{3}{c|}{\textbf{Basketball}} & \multicolumn{3}{c|}{\textbf{Boxes}} & \multicolumn{3}{c|}{\textbf{Football}} & \multicolumn{3}{c|}{\textbf{Juggle}} & \multicolumn{3}{c|}{\textbf{Softball}} & \multicolumn{3}{c|}{\textbf{Tennis}} & \multicolumn{3}{c}{\textbf{Mean}} \\ \cmidrule{2-22} 
\multicolumn{1}{c|}{}
 & \textbf{PSNR} & \textbf{SSIM} & \textbf{LPIPS} & \textbf{PSNR} & \textbf{SSIM} & \textbf{LPIPS} & \textbf{PSNR} & \textbf{SSIM} & \textbf{LPIPS} & \textbf{PSNR} & \textbf{SSIM} & \textbf{LPIPS} & \textbf{PSNR} & \textbf{SSIM} & \textbf{LPIPS} & \textbf{PSNR} & \textbf{SSIM} & \textbf{LPIPS} & \textbf{PSNR} & \textbf{SSIM} & \textbf{LPIPS} \\ \midrule
 \textcolor{black}{HyperReel} \cite{Hyperreel}$^1$ & 18.88 & 0.687 & 0.543 & 18.71 & 0.659 & 0.556 & 19.10 & 0.690 & 0.533 & 19.72 & 0.702 & 0.507 & 19.17 & 0.702 & 0.525 & 19.35 & 0.692 & 0.525 & 19.15 & 0.688 & 0.532 \\
\textcolor{black}{K-Planes} \cite{K-planes}$^1$ & 17.38 & 0.666 & 0.577 & 18.70 & 0.683 & 0.571 & 18.78 & 0.704 & 0.536 & 19.05 & 0.760 & 0.579 & 19.07 & 0.750 & 0.559 & 17.37 & 0.675 & 0.591 & 18.39 & 0.706 & 0.569 \\ 
 4D-Gaussians \cite{4d-gaussian-splatting}$^2$ & 22.63 & 0.763 & 0.356 & 23.06 & 0.792 & 0.491 & 22.42 & 0.800 & 0.563 & 23.20 & 0.805 & 0.492 & 23.07 & 0.806 & 0.527 & 23.30 & 0.803 & 0.495 & 22.95 & 0.795 & 0.487 \\
STGaussian \cite{spacetime}$^2$ & 20.83 & 0.789 & 0.629 & 21.29 & 0.795 & 0.603 & 21.41 & 0.804 & 0.603 & 22.22 & 0.812 & 0.578 & 21.89 & 0.808 & 0.600 & 22.51 & 0.809 & 0.575 & 21.69 & 0.803 & 0.598 \\
\textcolor{black}{4DRotorGS} \cite{4d-rotor}$^2$ & 22.92 & 0.803 & 0.446 & 22.44 & 0.800 & 0.459 & 22.84 & 0.810 & 0.440 & 24.09 & 0.821 & 0.428 & 22.99 & 0.812 & 0.427 & 23.03 & 0.813 & 0.461 & 23.05 & 0.810 & 0.443 \\ 
4DGS \cite{4dgs}$^3$ & 21.32 & 0.773 & 0.367 & 21.44 & 0.772 & 0.306 & 21.83 & 0.773 & 0.303 & 22.07 & 0.779 & 0.299 & 21.23 & 0.783 & 0.332 & 21.23 & 0.777 & 0.324 & 21.52 & 0.776 & 0.322 \\
Deformable-GS \cite{Deformable-Gaussians}$^3$ & 20.29 & 0.711 & 0.428 & 21.24 & 0.724 & 0.338 & 20.75 & 0.709 & 0.401 & 20.68 & 0.744 & 0.380 & 20.85 & 0.743 & 0.393 & 20.47 & 0.733 & 0.421 & 20.71 & 0.727 & 0.393 \\ 
\textcolor{black}{3DGStream} \cite{3dgstream}$^4$ & 24.28 & 0.789 & 0.276 & 25.97 & 0.848 & 0.195 & 26.04 & 0.836 & 0.222 & 25.96 & 0.850 & 0.211 & 25.50 & 0.841 & 0.215 & 25.53 & 0.844 & 0.213 & 25.55 & 0.835 & 0.222 \\
3DGS-O \cite{3DGS}$^4$ & \cellcolor{tablethird} 28.26 & \cellcolor{tablethird} 0.889 & \cellcolor{tablethird} 0.143 & \cellcolor{tablesecond}  28.34 & \cellcolor{tablethird} 0.893 & \cellcolor{tablethird} 0.139 & \cellcolor{tablethird} 28.00 & \cellcolor{tablethird} 0.894 & \cellcolor{tablesecond} 0.147 & \cellcolor{tablethird} 28.66 & \cellcolor{tablethird} 0.902 & \cellcolor{tablethird} 0.131 & \cellcolor{tablethird} 28.54 & \cellcolor{tablethird} 0.898 & \cellcolor{tablethird} 0.139 & \cellcolor{tablethird} 26.72 & \cellcolor{tablethird} 0.895 & \cellcolor{tablethird} 0.140 & \cellcolor{tablethird} 28.09 & \cellcolor{tablethird} 0.895 & \cellcolor{tablethird} 0.140 \\
D-3DGS \cite{d3dg}$^4$ & \cellcolor{tablesecond} 28.50 & \cellcolor{tablesecond}  0.892 & \cellcolor{tablesecond}  0.135 & \cellcolor{tablethird} 27.17 & \cellcolor{tablesecond}  0.896 & \cellcolor{tablesecond}  0.137 & \cellcolor{tablesecond}  28.10 & \cellcolor{tablesecond}  0.897 & \cellcolor{tablethird} 0.148 & \cellcolor{tablesecond} 28.77 & \cellcolor{tablesecond} 0.905 & \cellcolor{tablesecond} 0.127 & \cellcolor{tablesecond} 28.62 & \cellcolor{tablesecond} 0.902 & \cellcolor{tablesecond} 0.135 & \cellcolor{tablesecond} 27.70 & \cellcolor{tablesecond} 0.903 & \cellcolor{tablesecond} 0.125 & \cellcolor{tablesecond} 28.14 & \cellcolor{tablesecond} 0.899 & \cellcolor{tablesecond} 0.134 \\
Ours $^4$ & \cellcolor{tablefirst}  29.79 & \cellcolor{tablefirst} 0.907 & \cellcolor{tablefirst} 0.111 & \cellcolor{tablefirst} 29.77 & \cellcolor{tablefirst} 0.907 & \cellcolor{tablefirst} 0.108 & \cellcolor{tablefirst} 29.90 & \cellcolor{tablefirst} 0.912 & \cellcolor{tablefirst} 0.112 & \cellcolor{tablefirst} 30.29 & \cellcolor{tablefirst} 0.916 & \cellcolor{tablefirst} 0.105 & \cellcolor{tablefirst} 29.69 & \cellcolor{tablefirst} 0.911 & \cellcolor{tablefirst} 0.113 & \cellcolor{tablefirst} 29.53 & \cellcolor{tablefirst} 0.915 & \cellcolor{tablefirst} 0.108 & \cellcolor{tablefirst} 29.83 & \cellcolor{tablefirst} 0.911 & \cellcolor{tablefirst} 0.109 \\ \bottomrule
\end{tabular}
}
\\
\vspace{0.2em} 
{\footnotesize
\raggedright
$^1$NeRF-based methods. $^2$Temporal GS methods. $^3$Deformation-field GS methods. $^4$Per-timestamp training GS methods. 
}
\end{table*}

\begin{table*}[!t]
\scriptsize
\centering
\caption{\textbf{Per-scene quantitative comparisons of novel view synthesis on self-captured dataset.}}
\label{ous_render_table}
\renewcommand\arraystretch{1.2}
\setlength{\tabcolsep}{0.5mm}
{\begin{tabular}{l|ccc|ccc|ccc|ccc|ccc|ccc|ccc}
\toprule
\multirow{2}{*}{\textbf{Methods}} & \multicolumn{3}{c|}{\textbf{apple\_move}} & \multicolumn{3}{c|}{\textbf{sponge\_bowl}} & \multicolumn{3}{c|}{\textbf{orange\_bowl}} & \multicolumn{3}{c|}{\textbf{app.\_pan\_ora.\_bowl}} & \multicolumn{3}{c|}{\textbf{cloth\_pan}} & \multicolumn{3}{c|}{\textbf{building\_block}} & \multicolumn{3}{c}{\textbf{Mean}} \\ \cmidrule{2-22} 
 & \textbf{PSNR} & \textbf{SSIM} & \textbf{LPIPS} & \textbf{PSNR} & \textbf{SSIM} & \textbf{LPIPS} & \textbf{PSNR} & \textbf{SSIM} & \textbf{LPIPS} & \textbf{PSNR} & \textbf{SSIM} & \textbf{LPIPS} & \textbf{PSNR} & \textbf{SSIM} & \textbf{LPIPS} & \textbf{PSNR} & \textbf{SSIM} & \textbf{LPIPS} & \textbf{PSNR} & \textbf{SSIM} & \textbf{LPIPS} \\ \midrule
3DGS-O \cite{3DGS} & \cellcolor{tablesecond} 14.78 & \cellcolor{tablesecond} 0.558 & \cellcolor{tablesecond} 0.608 & \cellcolor{tablesecond} 15.05 & \cellcolor{tablethird} 0.551 & \cellcolor{tablesecond} 0.627 & \cellcolor{tablesecond} 14.76 & \cellcolor{tablesecond} 0.533 & \cellcolor{tablesecond} 0.641 & \cellcolor{tablethird} 14.22 & \cellcolor{tablethird} 0.520 & \cellcolor{tablesecond} 0.640 & \cellcolor{tablesecond} 15.40 & \cellcolor{tablesecond} 0.583 & \cellcolor{tablesecond} 0.595 & \cellcolor{tablesecond} 15.45 & \cellcolor{tablethird} 0.547 & \cellcolor{tablethird} 0.638 & \cellcolor{tablesecond} 14.94 & \cellcolor{tablethird} 0.548 & \cellcolor{tablesecond} 0.625 \\
D-3DGS \cite{d3dg} & \cellcolor{tablethird} 14.68 & \cellcolor{tablethird} 0.549 & \cellcolor{tablethird} 0.635 & \cellcolor{tablethird} 14.92 & \cellcolor{tablesecond} 0.561 & \cellcolor{tablethird} 0.636 & \cellcolor{tablethird} 13.98 & \cellcolor{tablethird} 0.529 & \cellcolor{tablethird} 0.682 & \cellcolor{tablesecond} 14.23 & \cellcolor{tablesecond} 0.531 & \cellcolor{tablesecond} 0.640 & \cellcolor{tablethird} 14.82 & \cellcolor{tablethird} 0.573 & \cellcolor{tablethird} 0.614 & \cellcolor{tablethird} 15.37 & \cellcolor{tablesecond} 0.560 &\cellcolor{tablesecond}  0.637 & \cellcolor{tablethird} 14.67 & \cellcolor{tablesecond} 0.550 & \cellcolor{tablethird} 0.641 \\
Ours & \cellcolor{tablefirst} 16.51 & \cellcolor{tablefirst} 0.596 & \cellcolor{tablefirst} 0.562 & \cellcolor{tablefirst} 16.45 & \cellcolor{tablefirst} 0.580 & \cellcolor{tablefirst} 0.585 & \cellcolor{tablefirst} 16.59 & \cellcolor{tablefirst} 0.592 & \cellcolor{tablefirst} 0.576 & \cellcolor{tablefirst} 15.71 & \cellcolor{tablefirst} 0.557 & \cellcolor{tablefirst} 0.590 & \cellcolor{tablefirst} 16.48 & \cellcolor{tablefirst} 0.612 & \cellcolor{tablefirst} 0.567 & \cellcolor{tablefirst} 16.26 & \cellcolor{tablefirst} 0.571 & \cellcolor{tablefirst} 0.610 & \cellcolor{tablefirst} 16.33 & \cellcolor{tablefirst} 0.585 & \cellcolor{tablefirst} 0.582 \\  \bottomrule
\end{tabular}
}
\end{table*}

\begin{table}[!t]
\scriptsize
\centering
\caption{\textbf{Per-scene quantitative comparisons of novel view synthesis on  ParticleNeRF  dataset \cite{Particlenerf}.}}
\label{part_table}
\renewcommand\arraystretch{1.2}
\setlength{\tabcolsep}{0.4mm}
{\begin{tabular}{l|ccc|ccc|ccc}
\toprule
\multicolumn{1}{c|}{\multirow{2}{*}{\textbf{Methods}}} & \multicolumn{3}{c|}{\textbf{Robot\_task}} & \multicolumn{3}{c|}{\textbf{Pendulums}} & \multicolumn{3}{c}{\textbf{Mean}} \\ \cmidrule{2-10} 
\multicolumn{1}{c|}{} & \textbf{PSNR} & \textbf{SSIM} & \textbf{LPIPS} & \textbf{PSNR} & \textbf{SSIM} & \textbf{LPIPS} & \textbf{PSNR} & \textbf{SSIM} & \textbf{LPIPS} \\ \midrule
3DGS-O \cite{3DGS}& \cellcolor{tablesecond} 30.01 & \cellcolor{tablefirst}  0.973 & \cellcolor{tablethird} 0.057 & \cellcolor{tablethird} 29.98 & \cellcolor{tablethird} 0.975 & \cellcolor{tablethird} 0.042 & \cellcolor{tablesecond} 29.99
 & \cellcolor{tablesecond} 0.974 & \cellcolor{tablethird} 0.049 \\
D-3DGS \cite{d3dg} & \cellcolor{tablethird} 29.35 & \cellcolor{tablethird} 0.971 & \cellcolor{tablesecond} 0.055 & \cellcolor{tablesecond} 30.07 & \cellcolor{tablesecond} 0.976 & \cellcolor{tablesecond} 0.037 & \cellcolor{tablethird} 29.71 & \cellcolor{tablesecond} 0.974 & \cellcolor{tablesecond} 0.046 \\
Ours & \cellcolor{tablefirst} 30.77 & \cellcolor{tablefirst} 0.973 & \cellcolor{tablefirst} 0.053 & \cellcolor{tablefirst} 33.34 & \cellcolor{tablefirst} 0.985 & \cellcolor{tablefirst} 0.031 & \cellcolor{tablefirst} 32.05 & \cellcolor{tablefirst} 0.979 & \cellcolor{tablefirst} 0.042 \\ \bottomrule
\end{tabular}
}
\end{table}

\vspace{1em} \noindent \textbf{Real-world Dataset.}
The PanopticSports dataset \cite{d3dg} serves as the primary resource for evaluating the accuracy of rendering and tracking. This dataset comprises six human motion sequences, with 27 cameras designated for training and 4 for testing. Ground truth trajectories are extracted using OpenPose \cite{openpose} to recognize human skeletons, focusing on the wrists and elbows of each individual as the tracked 3D points.

\vspace{1em} \noindent \textbf{Synthetic Dataset.}
Two challenging scenes from the ParticleNeRF dataset \cite{Particlenerf} are employed as the synthetic dataset, featuring 20 training cameras and 10 testing cameras. 
To simulate larger inter-frame motions, we select one frame every five frames from the dataset.

\vspace{1em} \noindent \textbf{Self-Captured Dataset.}
We develop a real-world robotic arm platform equipped with a multi-camera acquisition system. As shown in Fig. \ref{self_captured}, the setup includes 5 cameras used for training (4 ORBBEC Femto Bolt and 1 Kinect Azure) and 1 Kinect Azure camera designated for testing. \textcolor{black}{Different camera types demonstrate the algorithm's generality across devices, with all cameras synchronized across multiple viewpoints using ROS's synchronization mechanism.
The robotic manipulator Agile Diana executes randomized remote-controlled grasping operations in a tabletop environment, generating three short and three long motion sequences.}
The manipulated objects include both rigid items, such as apples and oranges, and flexible ones, such as sponges and towels, each labeled with an ArUco marker to provide ground-truth pose information.

\vspace{1em} \noindent \textbf{Evaluation Metrics.}
\textcolor{black}{Standard metrics such as \underline{PSNR}, \underline{SSIM}, and \underline{LPIPS} are utilized for NVS evaluation. For 3D long-term point tracking, we adopt the metrics median trajectory error (\underline{MTE}), position accuracy (\underline{Acc}), and survival rate (\underline{Surv}) as outlined in \cite{d3dg}.} Here, Acc is calculated at thresholds of 1, 2, 4, 8, and 16 cm, while Surv measures the trajectories that remain within 50 cm of the ground truth.

\begin{figure}[!t]\centering
	\includegraphics[width=7.5cm]{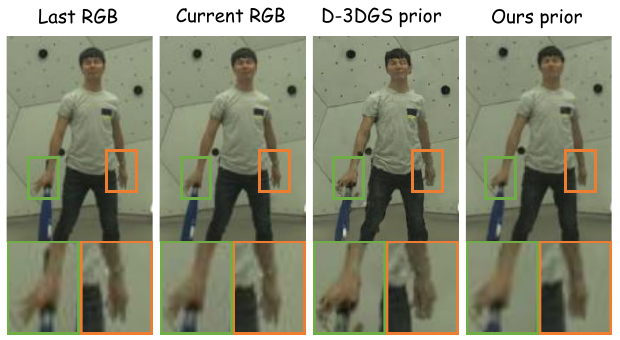}
	\caption{ \textbf{Qualitative comparisons of prior motion.} D-3DGS \cite{d3dg} tends to produce aberrations during irregular motion. In contrast, PaMoSplat achieves results nearly identical to current observations even before the Gaussian Splatting optimization pipeline.} 
	\label{prior_motion} 
\end{figure}

\subsection{Results} \label{results}

\noindent \textbf{Rendering Results.}
\textcolor{black}{As summarized in Tab. \ref{cmu_table}, PaMoSplat achieves the best rendering performance across all scenes in the PanopticSports dataset compared to other baselines. Qualitative comparisons are presented in Fig. \ref{render_main}. 
The NeRF-based approaches \cite{Hyperreel, K-planes} exhibit comparatively weaker performance due to multiple inherent limitations: their implicit reconstruction paradigm with sparse camera views frequently leads to overfitting on training viewpoints; the wall-mounted camera configuration in this dataset makes the system particularly sensitive to slight geometric deviations, often resulting in artifacts appearing in front of the cameras; and their implicit representations demand denser information encoding than explicit 3DGS methods, resulting in more challenging optimization.}

For the 3DGS-based methods, strategies that rely on extending the temporal dimension of Gaussians \cite{4d-gaussian-splatting, spacetime, 4d-rotor} or using deformation fields \cite{4dgs, Deformable-Gaussians} struggle to generalize to scenes with large motion and sparse viewpoints because the high complexity of the Gaussian motion patterns increases the difficulty of spatiotemporal fitting. 
Comparatively, the per-timestamp iterative strategy achieves higher rendering accuracy overall. Among them, 3DGStream \cite{3dgstream} yields lower metrics, mainly because it indirectly models Gaussian translations and rotations via a Neural Transformation Cache, which, despite enhancing the compactness of the model, compromises accuracy in the presence of large inter-frame motion.
{Although D-3DGS \cite{d3dg} incorporates physical priors, its loose rigidity constraints lack robustness against irregular motion patterns, compromising the fidelity of detailed rendering. In contrast, PaMoSplat explicitly models part-level deformation driven by optical flow, thereby preserving fine-grained structural details.}
Fig. \ref{prior_motion} compares the Gaussian-level motion prediction of D-3DGS with the part-level motion prediction of PaMoSplat, demonstrating the superior robustness of PaMoSplat.

\begin{table*}[!t]
\scriptsize
\centering
\caption{\textbf{Per-scene quantitative comparisons of 3D tracking on PanopticSports  dataset \cite{d3dg}.}
Our results achieve the highest metrics (\textbf{MTE [cm] $\downarrow$, Acc $\uparrow$, Surv $\uparrow$}) across most scenes.
}
\label{cmu_traj_table}
\renewcommand\arraystretch{1.2}
\setlength{\tabcolsep}{0.5mm}
{\begin{tabular}{l|ccc|ccc|ccc|ccc|ccc|ccc|ccc}
\toprule
\multirow{2}{*}{\textbf{Methods}} & \multicolumn{3}{c|}{\textbf{Basketball}} & \multicolumn{3}{c|}{\textbf{Boxes}} & \multicolumn{3}{c|}{\textbf{Football}} & \multicolumn{3}{c|}{\textbf{Juggle}} & \multicolumn{3}{c|}{\textbf{Softball}} & \multicolumn{3}{c|}{\textbf{Tennis}} & \multicolumn{3}{c}{\textbf{Mean}} \\ \cmidrule{2-22} 
 & \textbf{MTE} & \textbf{Acc} & \textbf{Surv} & \textbf{MTE} & \textbf{Acc} & \textbf{Surv} & \textbf{MTE} & \textbf{Acc} & \textbf{Surv} & \textbf{MTE} & \textbf{Acc} & \textbf{Surv} & \textbf{MTE} & \textbf{Acc} & \textbf{Surv} & \textbf{MTE} & \textbf{Acc} & \textbf{Surv} & \textbf{MTE} & \textbf{Acc} & \textbf{Surv} \\ \midrule
3DGS-O \cite{3DGS} & \cellcolor{tablethird} 29.63 & \cellcolor{tablethird} 10.32 & \cellcolor{tablethird} 26.48 & \cellcolor{tablethird} 29.32 & \cellcolor{tablethird} 07.50 & \cellcolor{tablethird} 24.83 & \cellcolor{tablethird} 28.88 & \cellcolor{tablethird} 13.10 & \cellcolor{tablethird} 31.58 & \cellcolor{tablethird} 18.79 & \cellcolor{tablethird} 16.87 & \cellcolor{tablethird} 43.83 & \cellcolor{tablethird} 21.76 & \cellcolor{tablethird} 16.71 & \cellcolor{tablethird} 36.22 & \cellcolor{tablethird} 114.74 & \cellcolor{tablethird} 07.69 & \cellcolor{tablethird} 15.56 & \cellcolor{tablethird} 40.52 & \cellcolor{tablethird} 29.75 & \cellcolor{tablethird} 12.03 \\
D-3DGS \cite{d3dg} &  \cellcolor{tablesecond} 04.88 & \cellcolor{tablesecond} 54.63 & \cellcolor{tablesecond} 93.81 & \cellcolor{tablesecond} 01.66 &  \cellcolor{tablefirst} 79.73 &  \cellcolor{tablefirst} 100.00 &\cellcolor{tablesecond}  05.02 & \cellcolor{tablesecond} 59.95 & \cellcolor{tablesecond} 93.75 & \cellcolor{tablesecond} 03.59 & \cellcolor{tablesecond} 66.47 &  \cellcolor{tablefirst} 100.00 & \cellcolor{tablefirst}  03.37 &  \cellcolor{tablefirst} 55.87 &  \cellcolor{tablefirst} 92.67 & \cellcolor{tablesecond} 02.23 & \cellcolor{tablesecond} 68.31 &  \cellcolor{tablefirst} 100.00 & \cellcolor{tablesecond} 03.46 & \cellcolor{tablesecond} 96.70 & \cellcolor{tablesecond} 64.16 \\
Ours &  \cellcolor{tablefirst} 02.27 &  \cellcolor{tablefirst} 67.26 &  \cellcolor{tablefirst} 100.00 &  \cellcolor{tablefirst} 01.57 & \cellcolor{tablesecond} 76.17 &  \cellcolor{tablefirst} 100.00 &  \cellcolor{tablefirst} 02.75 &  \cellcolor{tablefirst} 67.32 &  \cellcolor{tablefirst} 97.00 &  \cellcolor{tablefirst} 02.92 &  \cellcolor{tablefirst} 72.07 &  \cellcolor{tablefirst} 100.00 & \cellcolor{tablesecond} 03.83 & \cellcolor{tablesecond} 51.96 & \cellcolor{tablesecond} 91.78 &  \cellcolor{tablefirst} 01.94 &  \cellcolor{tablefirst} 73.07 &  \cellcolor{tablefirst} 100.00 &  \cellcolor{tablefirst} 02.55 &  \cellcolor{tablefirst} 98.13 &  \cellcolor{tablefirst} 67.97 \\ \bottomrule
\end{tabular}
}
\end{table*}

\begin{table*}[!t]
\scriptsize
\centering
\caption{\textbf{Per-scene quantitative comparisons of tracking on self-captured dataset.}
PaMoSplat achieves robust tracking accuracy even in scenes with sparse views and irregular motion.}
\label{ous_traj_table}
\renewcommand\arraystretch{1.2}
\setlength{\tabcolsep}{0.8mm}
{\begin{tabular}{l|ccc|ccc|ccc|ccc|ccc|ccc}
\toprule
\multirow{2}{*}{\textbf{MethodS}} & \multicolumn{3}{c|}{\textbf{apple\_move}} & \multicolumn{3}{c|}{\textbf{sponge\_bowl}} & \multicolumn{3}{c|}{\textbf{orange\_bowl}} & \multicolumn{3}{c|}{\textbf{app.\_pan\_ora.\_bowl}} & \multicolumn{3}{c|}{\textbf{building\_block}} & \multicolumn{3}{c}{\textbf{Mean}} \\ \cmidrule{2-19} 
 & \textbf{MTE} & \textbf{Acc} & \textbf{Surv} & \textbf{MTE} & \textbf{Acc} & \textbf{Surv} & \textbf{MTE} & \textbf{Acc} & \textbf{Surv} & \textbf{MTE} & \textbf{Acc} & \textbf{Surv} & \textbf{MTE} & \textbf{Acc} & \textbf{Surv} & \textbf{MTE} & \textbf{Acc} & \textbf{Surv} \\ \midrule
3DGS-O \cite{3DGS} & \cellcolor{tablethird} 15.49 & \cellcolor{tablethird} 14.07 & \cellcolor{tablesecond} 48.15 & \cellcolor{tablesecond} 13.07 & \cellcolor{tablesecond} 20.67 & \cellcolor{tablesecond} 60.00 & \cellcolor{tablesecond} 09.92 & \cellcolor{tablesecond} 30.38 & \cellcolor{tablesecond} 53.85 & \cellcolor{tablethird} 04.07 & \cellcolor{tablethird} 46.89 & \cellcolor{tablethird} 68.24 & \cellcolor{tablethird} 09.50 & \cellcolor{tablethird} 43.94 & \cellcolor{tablethird} 69.70 & \cellcolor{tablesecond} 10.41 & \cellcolor{tablethird} 31.19 & \cellcolor{tablethird} 59.99 \\
D-3DGS \cite{d3dg} & \cellcolor{tablesecond} 15.40 & \cellcolor{tablesecond} 18.52 & \cellcolor{tablesecond} 48.15 & \cellcolor{tablethird} 15.76 & \cellcolor{tablethird} 17.33 & \cellcolor{tablethird} 43.33 & \cellcolor{tablethird} 13.97 & \cellcolor{tablethird} 30.00 & \cellcolor{tablethird} 44.23 & \cellcolor{tablesecond} 03.90 & \cellcolor{tablesecond} 58.51 & \cellcolor{tablesecond} 68.92 & \cellcolor{tablesecond} 05.07 & \cellcolor{tablesecond} 57.58 & \cellcolor{tablesecond} 100.00 & \cellcolor{tablethird} 10.82 & \cellcolor{tablesecond} 36.39 & \cellcolor{tablesecond} 60.93 \\
Ours & \cellcolor{tablefirst} 01.55 & \cellcolor{tablefirst} 77.04 & \cellcolor{tablefirst} 100.00 & \cellcolor{tablefirst} 02.21 & \cellcolor{tablefirst} 72.00 & \cellcolor{tablefirst} 100.00 & \cellcolor{tablefirst} 01.85 & \cellcolor{tablefirst} 76.92 & \cellcolor{tablefirst} 100.00 & \cellcolor{tablefirst} 03.42 & \cellcolor{tablefirst} 65.27 &\cellcolor{tablefirst}  77.03 & \cellcolor{tablefirst} 02.79 & \cellcolor{tablefirst} 66.06 & \cellcolor{tablefirst} 100.00 & \cellcolor{tablefirst} 02.36 & \cellcolor{tablefirst} 71.46 & \cellcolor{tablefirst} 95.41 \\\bottomrule
\end{tabular}
}
\end{table*}

In subsequent experiments, we compare only the two main baselines, 3DGS-O \cite{3DGS} and D-3DGS \cite{d3dg}, which share similar concepts and exhibit comparable accuracy to PaMoSplat.
Tab. \ref{ous_render_table} and \ref{part_table} present quantitative rendering comparisons against two primary baselines \cite{3DGS, d3dg} for both the ParticleNeRF dataset and our self-captured dataset, with Fig. \ref{render_main} displaying qualitative results.
For the ParticleNeRF dataset, 3DGS-O exhibits disordered Gaussian primitives in high-velocity regions, such as the tips of the pendulums, due to insufficient motion modeling. While D-3DGS maintains structural compactness, it fails to capture dynamic details like rapid light/shadow transitions. In contrast, PaMoSplat efficiently models motion through part-level deformation, preserving more computational resources for detail optimization.
{The self-captured dataset shows generally lower rendering metrics, attributed to three experimental constraints: insufficient multi-view coverage due to the limited number of cameras, color inconsistencies caused by varying imaging conditions, and black background artifacts resulting from incomplete observations. Under these challenging settings, both 3DGS-O and D-3DGS produce significant dynamic artifacts in novel view synthesis due to insufficient viewpoint supervision. In comparison, PaMoSplat demonstrates superior robustness and maintains temporal stability, benefiting from its part segmentation and the integration of optical flow cues.}

\begin{figure}[!t]\centering
	\includegraphics[width=8.3cm]{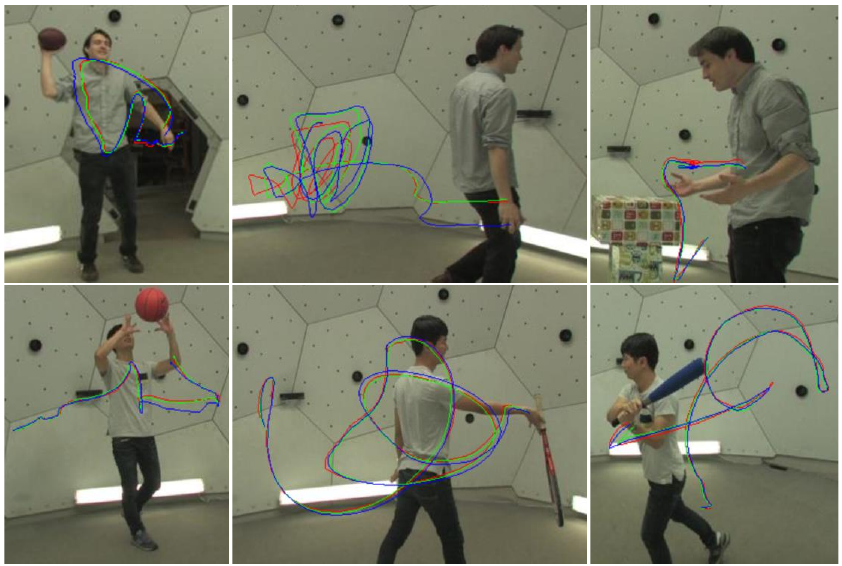}
	\caption{ \textbf{Qualitative comparisons of 3D trajectory (projected to 2D).} Our results (\textcolor{green}{green}) align more closely with the ground truth ({red}) values compared to D-3DGS \cite{d3dg} (\textcolor{blue}{blue}). Where ground truth is noisy, our results potentially offer greater accuracy.} 
	\label{traj} 
\end{figure}

\begin{figure}[!t]\centering
	\includegraphics[width=8.3cm]{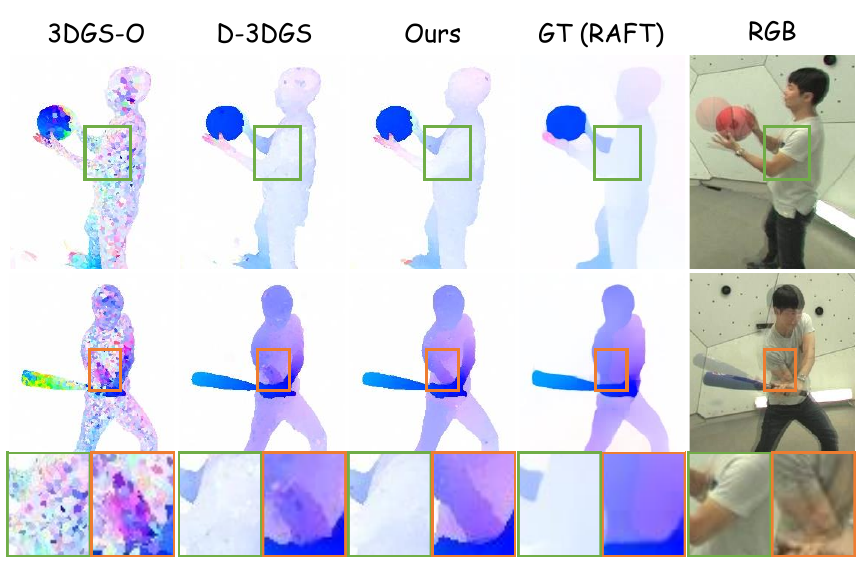}
	\caption{ \textcolor{black}{\textbf{Qualitative comparisons of calculated flow.} PaMoSplat produces less noise and reveals sharper boundaries.}}
	\label{flow} 
\end{figure}

\begin{figure}[!t]\centering
	\includegraphics[width=8.8cm]{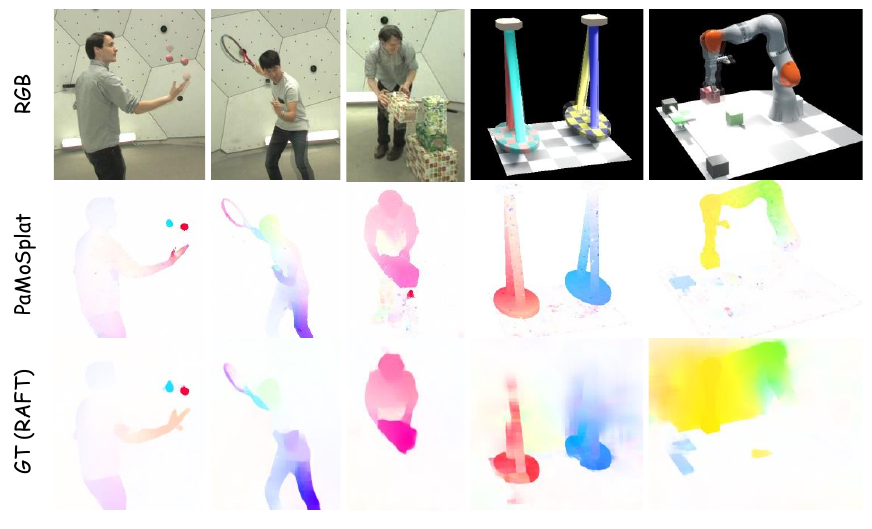}
	\caption{ \textcolor{black}{\textbf{More optical flow visualization.} PamoSplat even exceeds the performance of the front-end optical flow predictor RAFT.}}
	\label{flow_more} 
\end{figure}

\vspace{1em} \noindent \textbf{Tracking Results.}
The integration of Gaussian part and rigidity loss enables PaMoSplat to achieve effective 3D tracking. 
\textcolor{black}{Tab. \ref{cmu_traj_table} compares our 3D tracking performance to the main baselines 3DGS-O and D-3DGS, demonstrating our superior metrics in most scenes. Without introducing any physical constraints, 3DGS-O fails to ensure that Gaussians consistently represent specific spatial regions, leading to a near failure in tracking. 
As shown in Fig. \ref{traj}, our tracking trajectory is much closer to the ground truth compared to D-3DGS. This demonstrates that the motion and internal rigidity of PaMoSplat's Gaussian parts are more in line with real-world motion patterns than the undifferentiated local rigidity of D-3DGS. This is also corroborated by the optical flow results illustrated in Fig. \ref{flow}, which are obtained by projecting the 3D motion of the Gaussian primitives. PaMoSplat reduces noise and improves part consistency (e.g., the arm), and even exceeds the performance of the front-end optical flow predictor, RAFT \cite{Raft}.
We present additional optical flow results across various scenes in Fig. \ref{flow_more}.}

In the self-capture dataset, each object is labeled with an ArUco marker, allowing us to obtain the ground-truth 3D trajectory of each object along with depth information from the camera. For evaluation, we select 7 trajectories across 5 scenes  (the remaining scene is excluded due to challenges in extracting ground truth trajectories, such as occlusion and label blurring), as shown in Tab. \ref{ous_traj_table}. \textcolor{black}{In practice, tracking proves challenging in this scenario due to sparse viewpoint observations and the camera's close proximity to moving objects, which results in significant variations in 2D image-level rendering. PaMoSplat demonstrates stability in tracking highly dynamic objects, primarily due to its effective integration of optical flow information.} In contrast, the other two methods \cite{3DGS, d3dg} nearly fail in comparison.

\begin{figure}[!t]\centering
	\includegraphics[width=7cm]{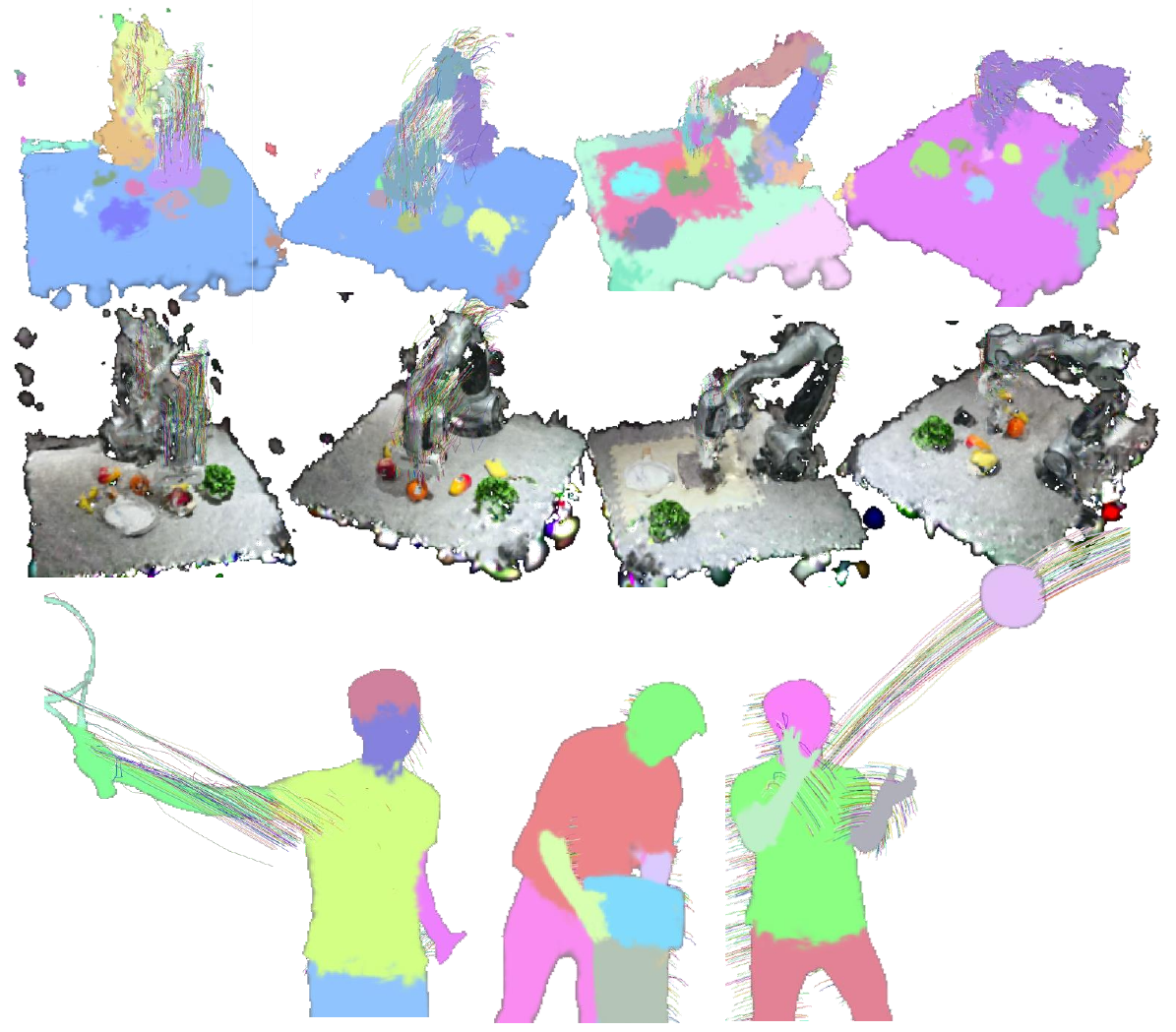}
	\caption{ \textbf{More Gaussian part visualizations.} For the self-captured dataset, we also provide Gaussian field color visualizations to facilitate cross-referencing.} 
	\label{part_all} 
\end{figure}

\vspace{1em} \noindent 
\textbf{More Gaussian Part Visualization.}
Fig. \ref{part_all} presents additional Gaussian part visualization results, demonstrating PaMoSplat’s capacity to accurately detect 3D parts with strong consistency in complex scenes during dynamic evolution. Although part segmentation granularity may vary (e.g., a robotic arm may be segmented as a single entity or divided by joints), the learnable rigidity $\mathcal{W}_{t}$ within each part provides robust adaptability to these variations.

\vspace{1em} \noindent 
\textbf{Results on the Mini-Movement Dataset.}
The above experiments demonstrate the significant advantages of PaMoSplat in handling large-scale, complex motion scenes. To further assess its performance, we test PaMoSplat on a mini-movement dataset: the Neural 3D Video dataset \cite{dynerf}, which contains multiple indoor sequences captured using 18–21 cameras. We select four scenes that are minimally influenced by outdoor lighting. Following standard practice, training and evaluation are conducted at half resolution, with the first camera designated for evaluation. 
\textcolor{black}{We select four nighttime scenes for evaluation.}

\begin{figure*}[!t]\centering
	\includegraphics[width=18cm]{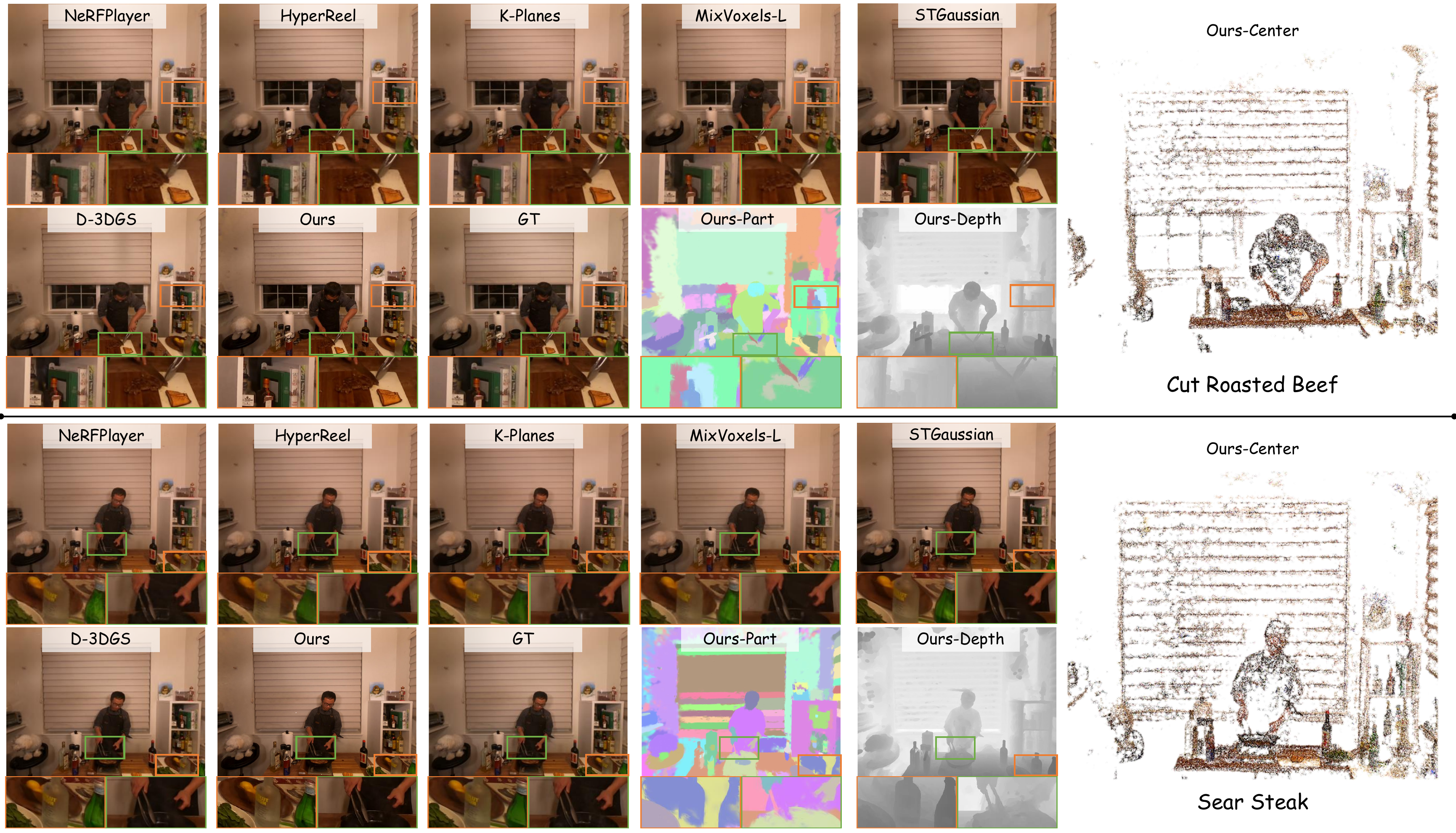}
	\caption{ \textbf{Qualitative comparison of novel view synthesis on Neural 3D Video dataset \cite{dynerf}.}  PaMoSplat demonstrates exceptional rendering fidelity, preserving fine details effectively.} 
	\label{neu3d_render_2} 
\end{figure*}

Qualitative and quantitative results are presented in Tab. \ref{n3d_table} and Fig. \ref{neu3d_render_2}. To ensure comparability, we report the PSNR metric, a common evaluation metric used in prior works such as \cite{4dgs} and \cite{spacetime}. Although the motion patterns in this dataset are relatively simple, the results effectively highlight the advanced performance of PaMoSplat in handling scenarios with less complex motion. This underscores its robustness and adaptability across various motion dynamics.
\textcolor{black}{Despite the training cameras being unilaterally distributed rather than $360^{\circ}$, PaMoSplat still produces superior part and depth images.}

\begin{table}[!t]
\scriptsize
\centering
\caption{\textbf{Per-scene quantitative comparisons of novel view synthesis in nighttime scenes on Neural 3D Video dataset \cite{dynerf}.} {Our results achieve the highest metric \textbf{PSNR}.}}
\label{n3d_table}
\renewcommand\arraystretch{1.2}
\setlength{\tabcolsep}{1.5mm}
{\begin{tabular}{l|cccc|c}
\toprule
\textbf{Methods} & \textbf{Cut R. B.} & \textbf{Sear S.} & \textbf{Cook S.} & \textbf{Flame S.} & \textbf{Mean} \\ \midrule
HexPlane \cite{hexplane} & 32.55 & 32.39 & 32.04 & 32.08 & 32.27 \\
NeRFPlayer \cite{Nerfplayer} & 29.35 & 29.13 & 30.56 & 31.39 & 30.11 \\
HyperReel \cite{Hyperreel} & 32.92 & 32.57 & 32.30 & 32.20 & 32.50 \\
K-Planes \cite{K-planes} & 31.82 & 32.52 & 32.60 & 32.38 & 32.33 \\
MixVoxels-L \cite{mixvoxel} & 32.40 & 32.10 & 32.25 & 31.83 & 32.15 \\
MixVoxels-X \cite{mixvoxel} & 32.63 & 32.33 & 32.31 & 32.10 & 32.34 \\
4DGS \cite{4dgs} & 32.90 & 32.49 & 32.46 & 32.51 & 32.59 \\
STGaussian \cite{spacetime} & \cellcolor[HTML]{FFE3B5}33.52 & \cellcolor[HTML]{FFE3B5}33.89 & \cellcolor[HTML]{FFE3B5}33.18 & \cellcolor[HTML]{FFE3B5}33.64 & \cellcolor[HTML]{FFE3B5}33.56 \\
\textcolor{black}{3DGStream} \cite{3dgstream} & \cellcolor[HTML]{FFFFDF}33.21 & 33.01 & \cellcolor[HTML]{FFC0CA}33.31 & \cellcolor[HTML]{FFC0CA}34.30 & \cellcolor[HTML]{FFFFDF}33.46 \\
D-3DGS \cite{d3dg} & 30.72 & \cellcolor[HTML]{FFFFDF}33.68 & 32.97 & 33.24 & 32.65 \\
Ours & \cellcolor[HTML]{FFC0CA}33.60 & \cellcolor[HTML]{FFC0CA}34.05 & \cellcolor[HTML]{FFFFDF}32.99 & \cellcolor[HTML]{FFFFDF}33.63 & \cellcolor[HTML]{FFC0CA}33.57 \\ \bottomrule
\end{tabular}
}
\end{table}

\begin{table}[!t]
\scriptsize
\centering
\caption{{\textbf{Comparison of computational cost.} Runtime differences across datasets arise from variations in image resolution.}}
\label{runtime}
\renewcommand\arraystretch{1.2}
\setlength{\tabcolsep}{1.0mm}
{
\begin{tabular}{clccc}
\toprule
\multicolumn{1}{l}{\textbf{}} & \textbf{Methods} & \textbf{PanopticSports} & \textbf{ParticleNeRF} & \textbf{Self-captured} \\ \midrule
 & D-3DGS \cite{d3dg} & \cellcolor[HTML]{FFE3B5}44.22±2.67 & \cellcolor[HTML]{FFE3B5}43.82±0.01 & \cellcolor[HTML]{FFE3B5}87.51±2.80 \\
\multirow{-2}{*}{\begin{tabular}[c]{@{}c@{}}Runtime per\\ Timestamp {[}s{]}\end{tabular}} & Ours & \cellcolor[HTML]{FFC0CA}26.03±3.36 & \cellcolor[HTML]{FFC0CA}30.24±0.46 & \cellcolor[HTML]{FFC0CA}71.43±7.76 \\ \midrule
 & D-3DGS \cite{d3dg} & \cellcolor[HTML]{FFE3B5}347.91±12.13 & \cellcolor[HTML]{FFE3B5}122.16±1.93 & \cellcolor[HTML]{FFE3B5}185.81±10.12 \\
\multirow{-2}{*}{\begin{tabular}[c]{@{}c@{}}Number   of\\      Gaussians {[}k{]}\end{tabular}} & Ours & \cellcolor[HTML]{FFC0CA}291.92±25.84 & \cellcolor[HTML]{FFC0CA}89.57±1.02 & \cellcolor[HTML]{FFC0CA}152.41±16.52 \\ \midrule
 & D-3DGS \cite{d3dg} & \cellcolor[HTML]{FFC0CA}3153±25 & \cellcolor[HTML]{FFC0CA}3162±11 & \cellcolor[HTML]{FFC0CA}3179±12 \\
\multirow{-2}{*}{\begin{tabular}[c]{@{}c@{}}GPU Memory \\ Usage {[}MB{]}\end{tabular}} & Ours & \cellcolor[HTML]{FFE3B5}4312±149 & \cellcolor[HTML]{FFE3B5}4232±19 & \cellcolor[HTML]{FFE3B5}4255±43 \\ \bottomrule
\end{tabular}
}
\end{table}

{
\subsection{Computational Cost Comparison} \label{RC}
While theoretically any method could achieve near-perfect results given sufficient training time and Gaussian ellipsoids, our comparison of PaMoSplat and D-3DGS (the primary competitive baseline) shows that PaMoSplat’s accuracy does not come at the expense of computational cost. We evaluate runtime per timestamp, number of Gaussians, and GPU memory usage, with results summarized in Tab. \ref{runtime}. Variations in Gaussian counts mainly stem from differences in densification settings. PaMoSplat attains higher accuracy with fewer Gaussians and faster processing, enabled by its Gaussian part prior motion and adaptive iteration strategy. In contrast, D-3DGS uses a fixed iteration schedule per timestamp, whose short-term local rigid constraints weaken under relative part motion, leading to longer per-iteration optimization. PaMoSplat does, however, require additional GPU memory to maintain Gaussian anchor distance $D$ and to support backpropagation of the part rigidity loss $\mathcal{L}_{part-rigid}$.
}

\begin{figure*}[!t]\centering
	\includegraphics[width=18cm]{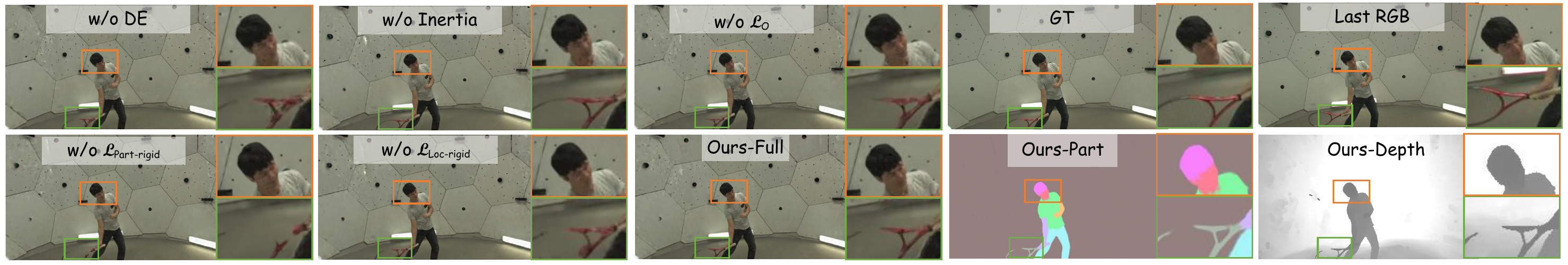}
	\caption{ {\textbf{Qualitative comparison of ablation studies.} Omitting any component results in reduced rendering and tracking accuracy.}} 
	\label{ablation} 
\end{figure*}

\begin{table}[!t]
\scriptsize
\centering
\caption{\textbf{Ablation results of proposed components.} Conducted on the \texttt{Tennis} scene from PanopticSports dataset \cite{d3dg}.}
\label{ablation_table}
\renewcommand\arraystretch{1.2}
\setlength{\tabcolsep}{1.2mm}
{\begin{tabular}{l|ccc|ccc}
\toprule
\multicolumn{1}{c|}{\multirow{2}{*}{\textbf{Description}}} & \multicolumn{3}{c|}{\textbf{Novel View Synthesis}} & \multicolumn{3}{c}{\textbf{3D Tracking}} \\ \cmidrule{2-7} 
\multicolumn{1}{c|}{} & \textbf{PSNR} & \textbf{SSIM} & \textbf{LPIPS} & \textbf{MTE} & \textbf{Acc} & \textbf{Surv} \\ \midrule
w/o DE & 28.64 & 0.914 & 0.110 & 05.47 & 81.78 & 42.67 \\
w/o Inertia & \cellcolor{tablethird} 29.34 & 0.914 & \cellcolor{tablethird} 0.108 & 02.68 & \cellcolor{tablethird} 98.66 & 63.95 \\
w/o $\mathcal{L}_{o}$ & 29.12 & \cellcolor{tablethird} 0.915 & \cellcolor{tablethird} 0.108 & \cellcolor{tablethird} 02.13 & \cellcolor{tablefirst} 100.00 & \cellcolor{tablesecond} 70.31 \\
w/o $\mathcal{L}_{Part-rigid}$ & \cellcolor{tablesecond} 29.51 & \cellcolor{tablefirst}0.918 & \cellcolor{tablefirst} 0.100 & \cellcolor{tablesecond} 01.98 & 97.15 & \cellcolor{tablethird} 65.12 \\
w/o $\mathcal{L}_{Loc-rigid}$ & 29.22 & \cellcolor{tablefirst}0.918 & \cellcolor{tablesecond} 0.103 & 33.56 & 19.56 & 07.56 \\
Ours-Full & \cellcolor{tablefirst} 29.53 & \cellcolor{tablethird} 0.915 & \cellcolor{tablethird} 0.108 & \cellcolor{tablefirst} 01.94 & \cellcolor{tablefirst} 100.00 & \cellcolor{tablefirst} 73.07 \\ \bottomrule
\end{tabular}
}
\end{table}

\subsection{Ablation Study} \label{AS}

To evaluate the effectiveness of proposed components, we conduct an ablation study in Tab. \ref{ablation_table} and Fig. \ref{ablation} using \texttt{Tennis} scene from the PanopticSports dataset \cite{d3dg}. 
Both `w/o DE' (Differential Evolution) and `w/o Inertia', two methods of eliminating prior motion, led to the largest drop in novel view synthesis metrics due to challenges in quickly fitting new timestamped scenes. They hinder the modeling of fast-moving objects, leading to artifacts, such as the distorted tennis bat. 
Omitting the flow-supervised loss (`w/o $\mathcal{L}_{o}$') made it difficult to focus optimization on dynamic regions, negatively impacting both rendering and tracking. The absence of rigidity losses (`w/o $\mathcal{L}_{Part-rigid}$' and `w/o $\mathcal{L}_{Loc-rigid}$') primarily reduced tracking accuracy. The removal of $\mathcal{L}_{Part-rigid}$ disrupts internal rigidity, causing overall part-level disarray, such as the incomplete tennis bat, while excluding $\mathcal{L}_{Loc-rigid}$ leads to subtle deformations in local structures, such as distorted faces. In contrast, the full PaMoSplat demonstrates superior performance.

\begin{figure*}[!t]\centering
	\includegraphics[width=17.8cm]{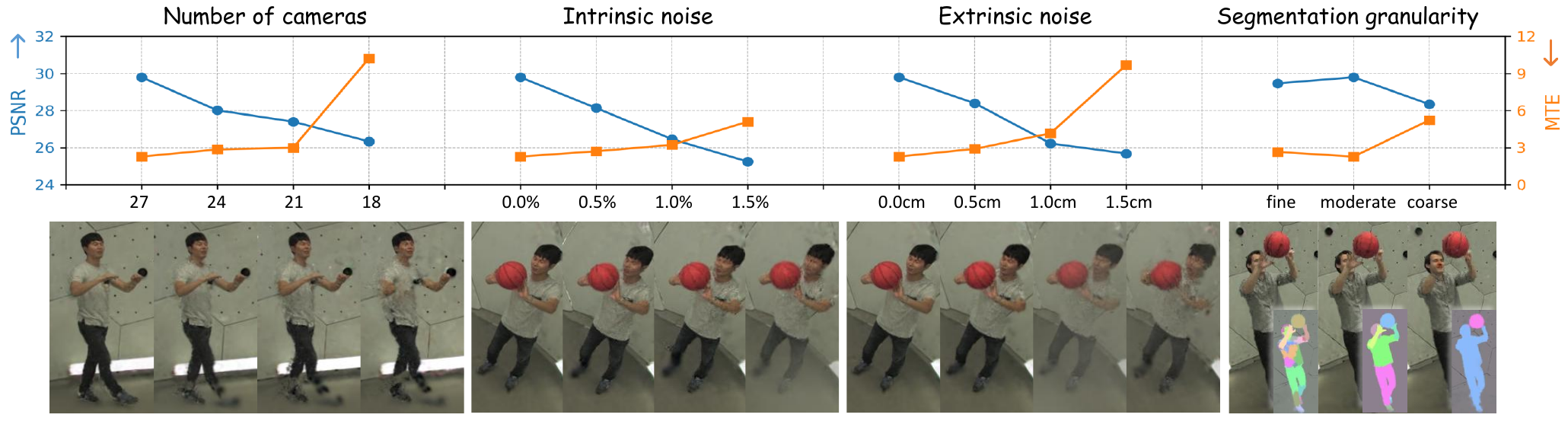}
	\caption{ \textcolor{black}{\textbf{Experiments on parameter influence.} Reducing the number of training cameras or introducing noise into camera parameters leads to degradation in rendering and tracking quality. While part segmentation granularity shows some robustness, performance deteriorates significantly under severe under-segmentation.}}
	\label{params} 
\end{figure*}

\begin{figure}[!t]\centering
	\includegraphics[width=8.8cm]{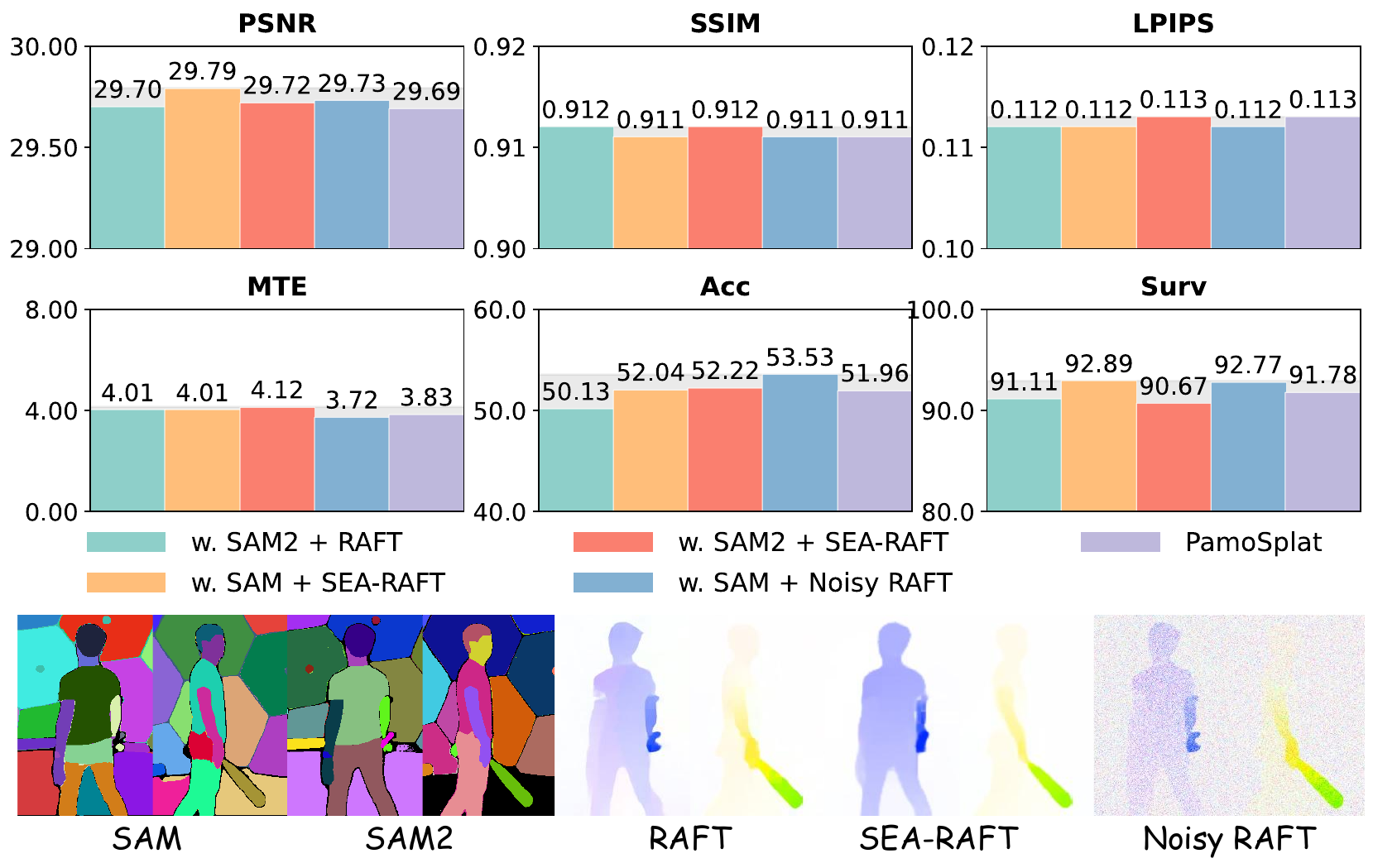}
        	\caption{ {\textbf{Model-agnostic study.} PamoSplat exhibits consistently stable performance across different models.}} 
	\label{new_model} 
\end{figure}

\begin{figure}[!t]\centering
	\includegraphics[width=7.3cm]{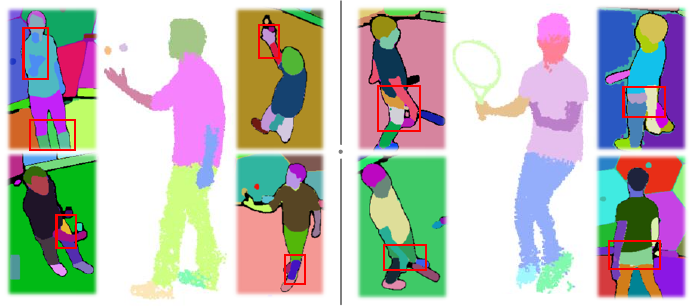}
        	\caption{ {\textbf{Robustness to front-end inaccurate segmentation.} Our graph clustering-based Gaussian part generation method demonstrates strong robustness to imperfect input.}} 
	\label{seg_robust} 
\end{figure}

{
\subsection{Model-Agnostic Study}
The PamoSplat pipeline relies on two primary frontend models: SAM \cite{sam} for 2D segmentation and RAFT \cite{Raft} for optical flow estimation. To assess their impact, we conduct a model-agnostic study on the \texttt{Softball} scene, evaluating alternative variants including SAM2 \cite{sam2}, SEA-RAFT \cite{sea_raft}, and an artificially degraded Noisy RAFT. 
As shown in Fig. \ref{new_model}, PamoSplat exhibits low sensitivity to the choice of frontend models. Although SAM2 and SEA-RAFT achieve slightly better intermediate segmentation or flow predictions, the final rendering and tracking metrics remain stable. For part segmentation, graph clustering improves robustness against inaccurate masks (as shown in Fig. \ref{seg_robust}).
For optical flow, the part-level formulation substantially enhances stability. Even under severely noisy flow inputs, the optimization reliably guides Gaussian parts toward plausible poses.
These results demonstrate that PamoSplat can be paired with any model providing similar functionality.
}

\subsection{\textcolor{black}{The Impact of parameters}} \label{IP}

\textcolor{black}{In this subsection, we investigate the effects of several key parameters on PaMoSplat, including the number of training cameras, the intrinsic and extrinsic camera parameters, and the granularity of Gaussian part segmentation. Fig. \ref{params} presents the PSNR metrics for rendering and MTE metrics for tracking, along with visual rendering results using \texttt{Basketball} scene from PanopticSports dataset \cite{d3dg}.}

\textcolor{black}{
\textbf{i) Number of cameras: }
As the number of cameras decreases, both rendering and tracking quality gradually decline, although a performance plateau is maintained in the early stages. Notably, when the number of cameras is reduced from 21 to 18, tracking accuracy degrades significantly, due to insufficient multi-view supervision for 3D part motion, e.g., the right arm disappears in the rendered image.}

\textcolor{black}{
\textbf{ii) Intrinsic noise: }
To simulate varying degrees of intrinsic parameter noise, we apply random perturbations to the focal length within a relative percentage range. As the intrinsic parameter noise increases, both rendering and tracking metrics degrade smoothly. The impact primarily manifests in the modeling of fine-grained details, e.g., increasingly distorted facial features.}

\textcolor{black}{
\textbf{iii) Extrinsic noise: }
Introducing noise into the extrinsic camera parameters yields similar trends. However, tracking performance is more sensitive to extrinsic noise, since perturbations in camera pose directly affect multi-view consistency and, consequently, the accuracy of Gaussian localization. At an extrinsic noise level of 1.5cm, the rendered human silhouette becomes notably blurred.}

\textbf{iv) Segmentation granularity:}
We also examine the influence of segmentation granularity. 
{We manually generated both fine-grained and coarse-grained segmentation.}
Fine-grained segmentation leads to minor performance drops, due to increased sensitivity to optical flow noise. In contrast, coarse-grained segmentation results in a notable decline in both metrics, as part-level motion priors are no longer accurately represented, which is consistent with the `w/o DE' ablation results discussed in the Sec. \ref{AS}.

\subsection{Further Applications} \label{FA}
\begin{figure}[!t]\centering
	\includegraphics[width=8.8cm]{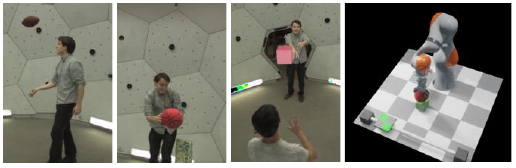}
        	\caption{ \textbf{Novel view synthesized by part-level scene editing.}} 
	\label{edit} 
\end{figure}

\textcolor{black}{PaMoSplat's novel part-aware representation fundamentally enables 4D scene manipulation capabilities at the part level, opening new possibilities for dynamic scene editing applications.
As illustrated in Fig. \ref{edit}, our framework allows selective replacement or modification of individual parts while perfectly preserving their original motion trajectories: modified parts move in complete synchronization with unaltered ones, maintaining physically plausible interactions.}
Additionally, the edited scene supports photorealistic image rendering from any view, with access to each part's motion trajectory. See \href{https://pamosplat.github.io/}{project website} for $360^{\circ}$ video of scene editing.

\subsection{\textcolor{black}{Limitations}} \label{limit}

\textcolor{black}{Although PaMoSplat demonstrates advantages in both 2D rendering and 3D tracking accuracy, it still has some limitations:}

\textcolor{black}{\textbf{i) Strict requirement for multi-view video input:} Certain key components of PaMoSplat, including Gaussian part segmentation and DE-induced prior motion, necessitate multi-view inputs, restricting its direct applicability to monocular video.}

\textcolor{black}{\textbf{ii) Inability to handle newly appearing objects:} PaMoSplat operates under the assumption that all objects exist at the initial timestamp, with subsequent frames requiring only Gaussian parameter updates rather than additions of new Gaussians. Consequently, the method cannot effectively model objects that appear at later timestamps.}

\begin{figure}[!t]\centering
	\includegraphics[width=8.8cm]{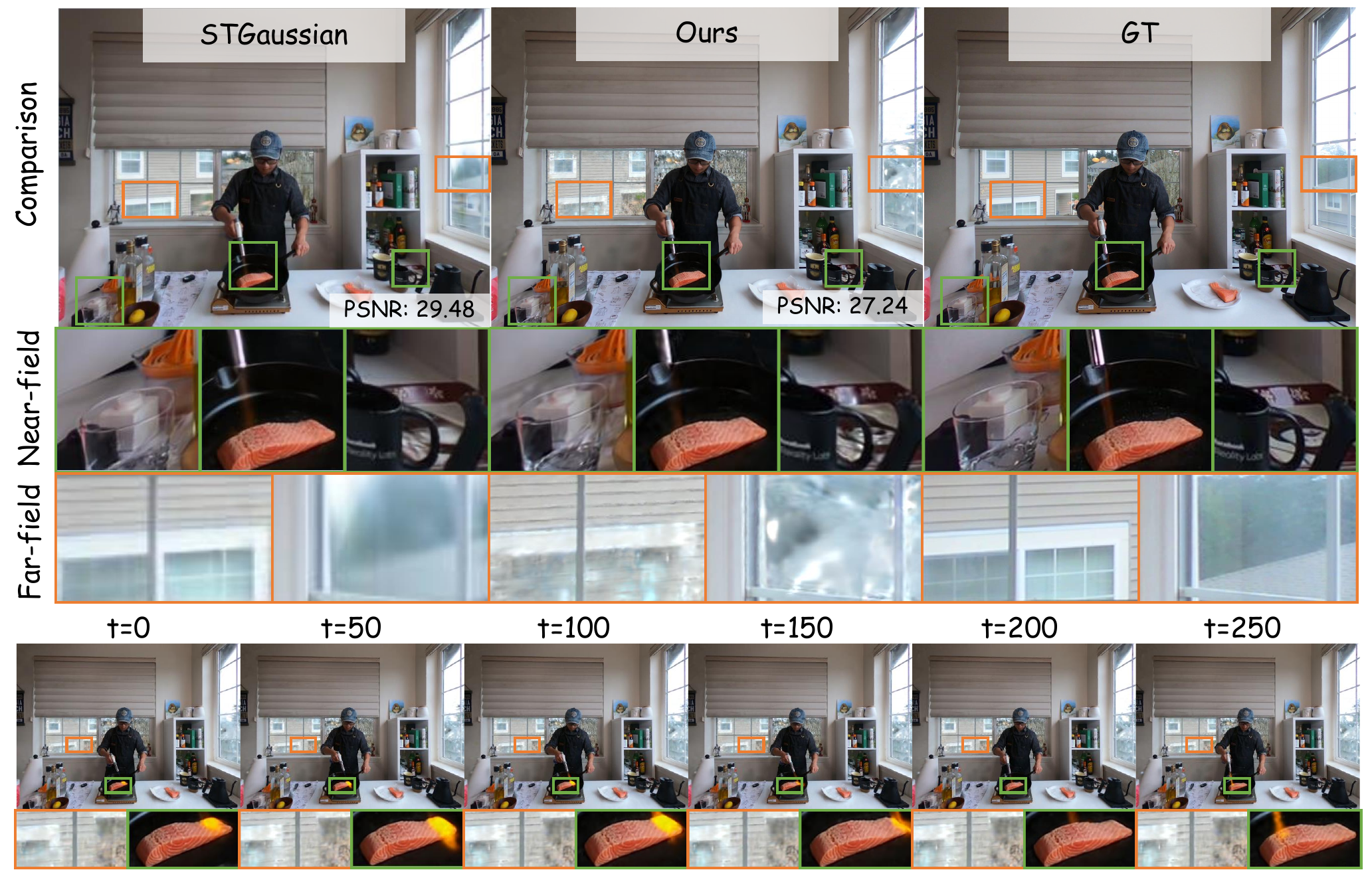}
        	\caption{ \textcolor{black}{\textbf{ Rendering results in daytime scene on Neural 3D Video dataset \cite{dynerf}.} While PaMoSplat is affected by the results of the 3DGS pipeline in initial timestamp, leading to noticeable distortions in the far-field from novel viewpoints, it still preserves higher rendering quality in the dynamic near-field.}}
	\label{daytime} 
\end{figure}

\textbf{iii) Temporal Dependency Limitation:}
{PaMoSplat inherits the fundamental constraint of per-timestamp training, where reconstruction quality at later timestamps is contingent upon earlier estimations.} This is particularly evident in our experiments with the Neural 3D Video dataset's daytime scene, as shown in Fig. \ref{daytime}, where novel view synthesis metrics are lower than those of STGaussian \cite{spacetime}. 
Upon closer examination of specific image regions, PaMoSplat's far-field rendering (highlighted in orange) is noticeably inferior to that of STGaussian. 
This discrepancy primarily arises from the Gaussian field $\mathcal{S}_0$ at the initial timestamp $t=0$ optimized by the 3DGS pipeline, which exhibits severe far-field artifacts, as illustrated in the bottom row of Fig. \ref{daytime}.
A similar trend is observed in D-3DGS \cite{d3dg} (PSNR = 26.92), which follows the same paradigm as PaMoSplat. Despite this, PaMoSplat maintains higher fidelity in the near field (highlighted in green), underscoring its advantage in modeling foreground dynamics. When the far-field regions (i.e., the two window areas) are excluded via a mask, PaMoSplat’s PSNR significantly improves from 27.24 to 31.16, corroborating the above analysis.



\section{Conclusion}
We present a novel part-aware motion-guided Gaussian splatting framework for dynamic scene reconstruction. In this framework, Gaussian parts serve as the fundamental units for deformation, with multi-view optical flow guiding their prior motion. 
{To further enhance dynamic modeling robustness, we introduce a flow-supervised rendering loss and a learnable internal rigidity loss. In addition, an adaptive iteration count mechanism maintains the computational efficiency of PaMoSplat. 
Experiments across diverse datasets show that our representation delivers superior rendering and tracking accuracy at competitive computational cost, while also enabling part-level downstream applications.}

PaMoSplat demonstrates the benefits of decomposable scene representations for both dynamic novel-view synthesis and dense persistent tracking. These capabilities naturally suggest several promising research directions:
{\textbf{1) Data Augmentation via Dynamic Gaussian Fields:}
As high-quality robotics data becomes increasingly critical for embodied intelligence, dynamic 3DGS fields offer an attractive solution for realistic and scalable data generation, leveraging their inherent strengths in photorealistic rendering and operational flexibility.}
\textbf{2) Real-Time Dynamic Gaussian Fields for Digital Twins:}
Further development could enable fully real-time dynamic Gaussian fields, creating digital twins that synchronize seamlessly with the physical world. Such systems would support applications ranging from online status monitoring to world model construction.

\bibliographystyle{IEEEtran}
\bibliography{TCSVT}

\begin{IEEEbiography}[{\includegraphics[width=1in,height=1.25in,clip,keepaspectratio]{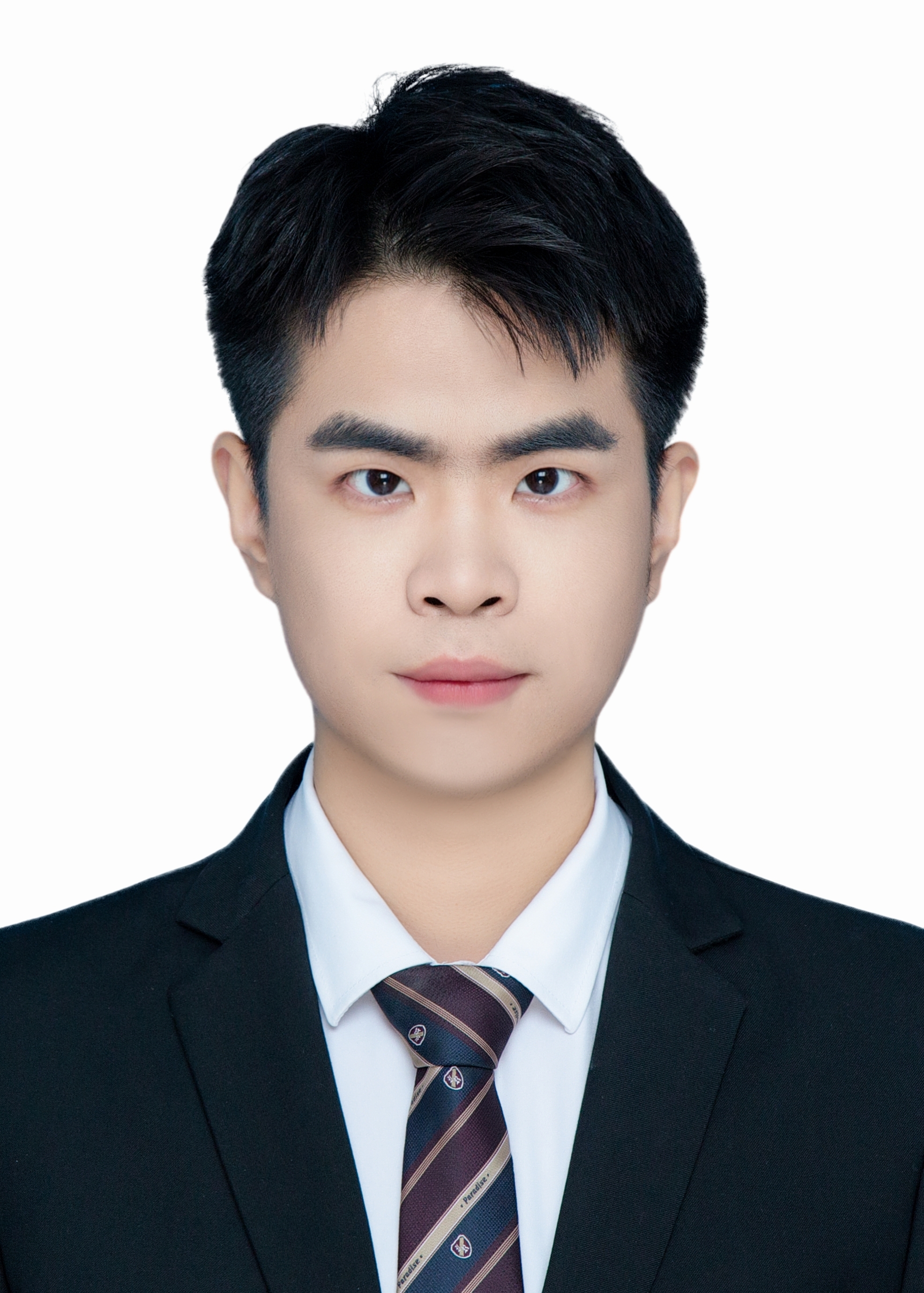}}]{Yinan Deng} (Student Member, IEEE) received the B.Eng. degree in automation from Beijing Institute of Technology, Beijing, China, in 2021. He is currently pursuing his Ph.D. degree in Control Science and Engineering at the School of Automation, Beijing Institute of Technology, China. His research interests include 3D vision, open vocabulary scene reconstruction, and embodied perception.
 \end{IEEEbiography}

\begin{IEEEbiography}[{\includegraphics[width=1in,height=1.25in,clip,keepaspectratio]{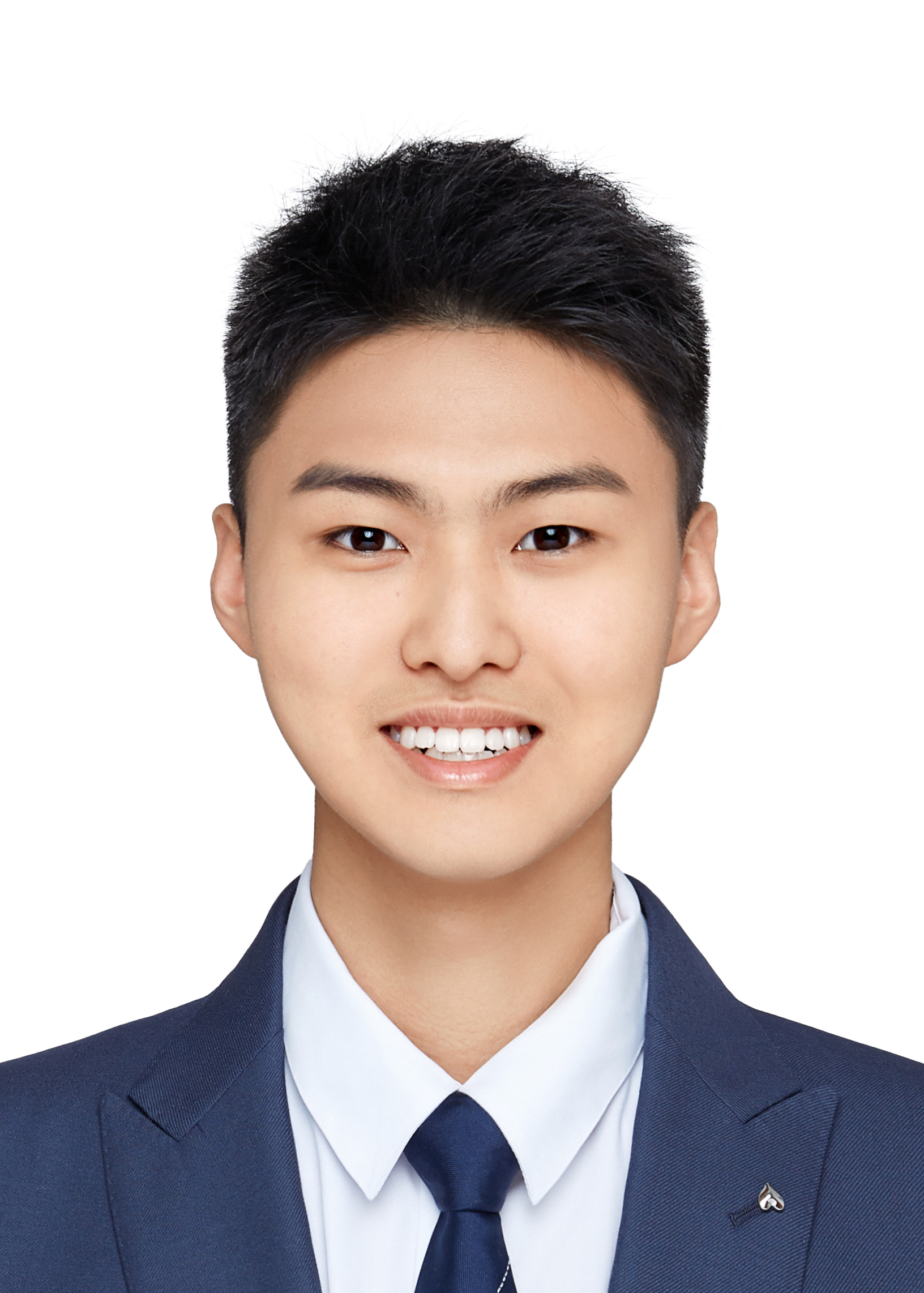}}]{Jianyu Dou} received the B.Eng. degree in automation from Beijing Institute of Technology, Beijing, China, in 2025. He has just become a PhD student in Control Science and Engineering at the School of Automation, Beijing Institute of Technology, China.  He has published a journal paper on multi-agent collaborative mapping, and his current research interests focus on data generation.
 \end{IEEEbiography}

\begin{IEEEbiography}[{\includegraphics[width=1in,height=1.25in,clip,keepaspectratio]{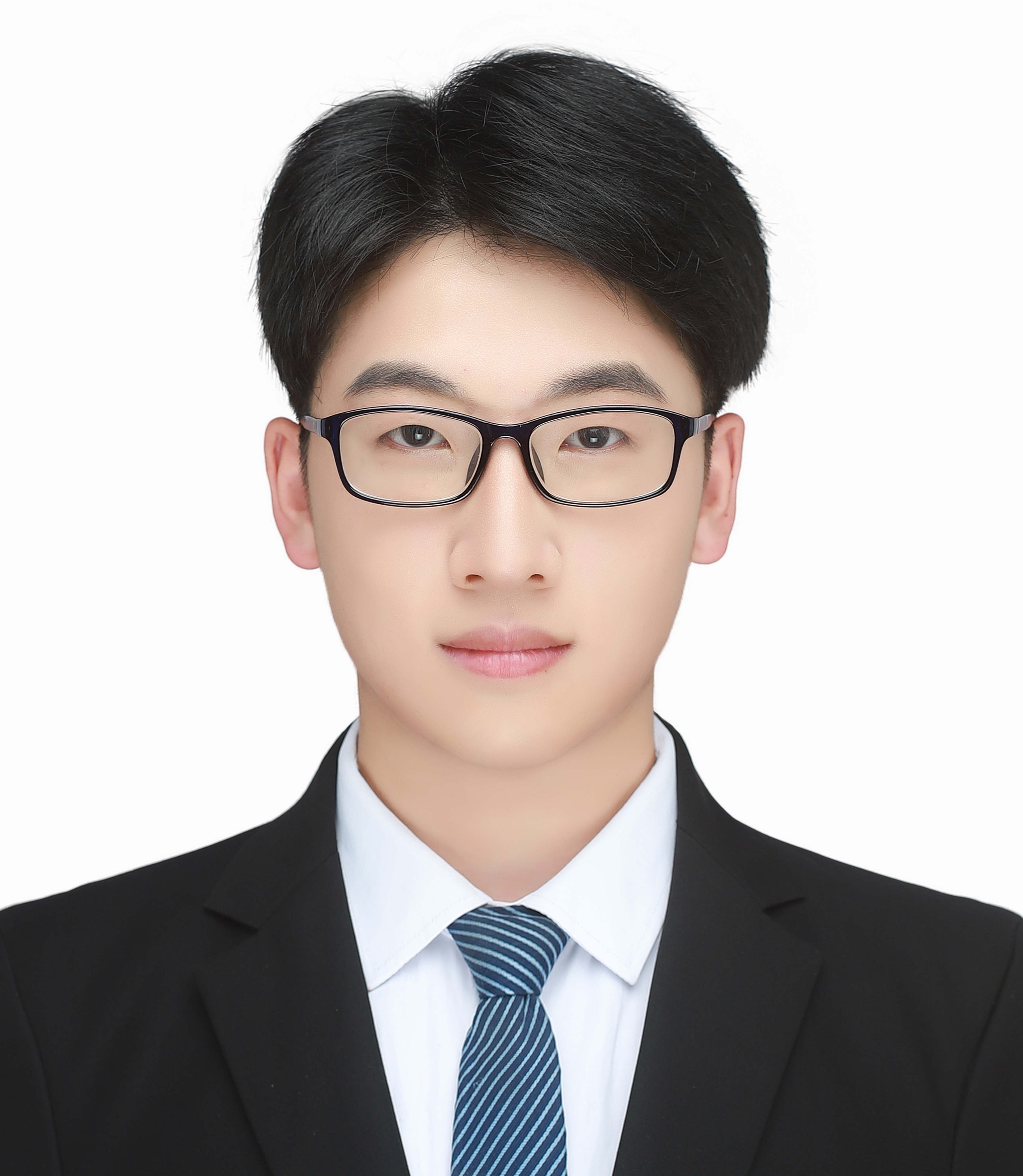}}]{Jiahui Wang} received the B.Eng. degree in automation from Central South University, Changsha, China, in 2023. He is currently working toward the Ph.D. degree in Control Science and Engineering for the third year with School of Automation, Beijing Institute of Technology, Beijing, China. His research interests include multi-modal mapping and language embedded representation for robotic application.
 \end{IEEEbiography}

\begin{IEEEbiography}[{\includegraphics[width=1in,height=1.25in,clip,keepaspectratio]{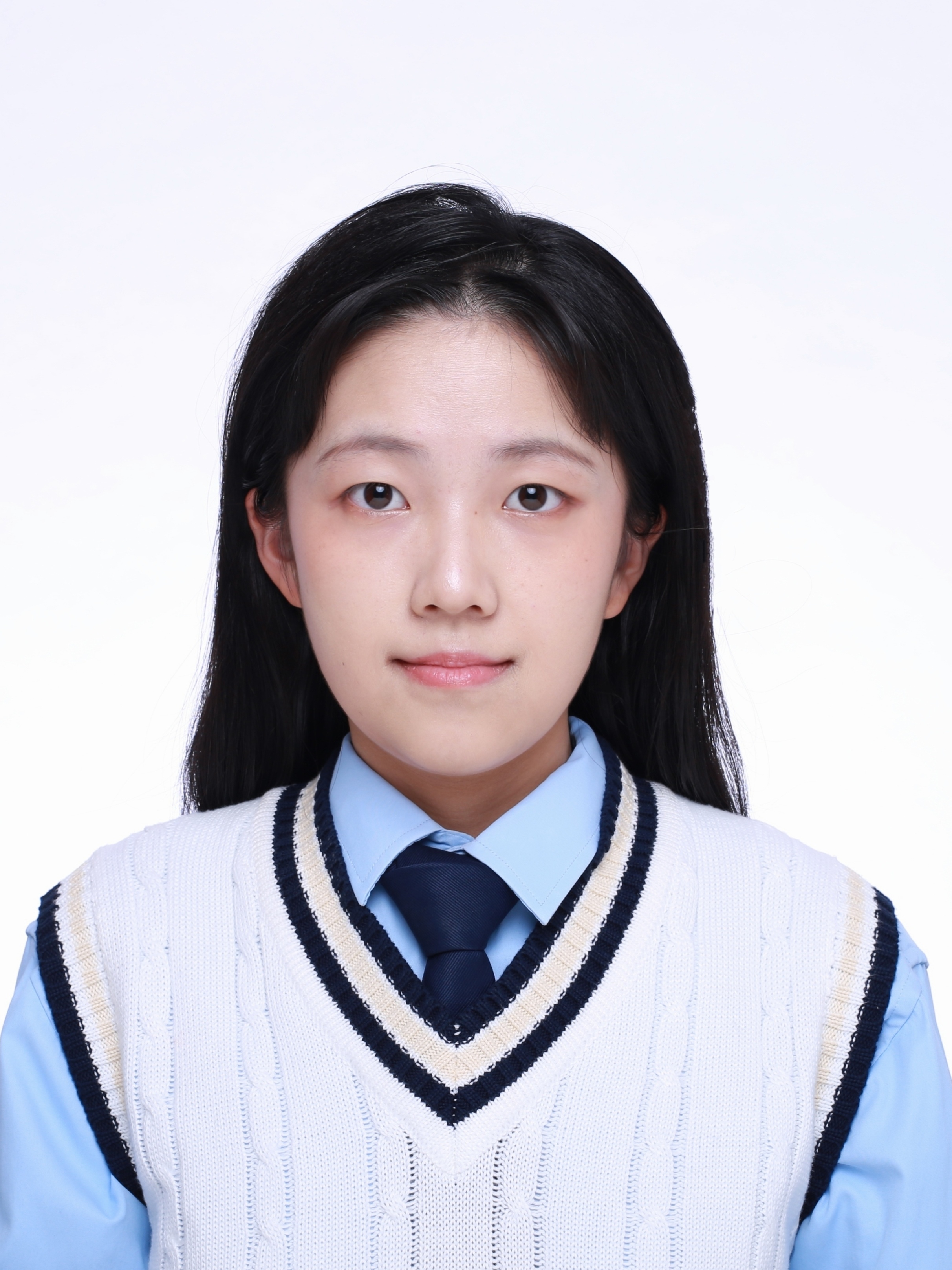}}]{Jingyu Zhao} received the B.Eng. degree in automation from the Beijing Institute of Technology, Beijing, China. She is currently pursuing a M.S. degree in control science and Engineering at the Beijing Institute of Technology. Her research interests focus on 3D scene reconstruction and understanding.
 \end{IEEEbiography}

\begin{IEEEbiography}[{\includegraphics[width=1in,height=1.25in,clip,keepaspectratio]{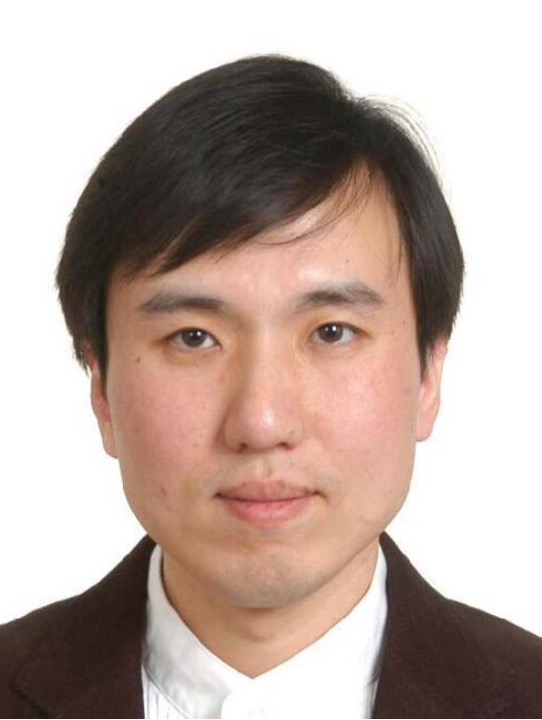}}]{Yi Yang} (Member, IEEE) received the B.Eng. degree in automation and the M.S. degree in control science and engineering from the Hebei University of Technology, Tianjin, China, in 2001 and 2004, respectively, and the Ph.D. degree in control science and engineering from the Beijing Institute of Technology, Beijing, China, in 2010. He is currently a Professor with the School of Automation, Beijing Institute of Technology.

His research interests include the areas of autonomous vehicles and robotics, with focus on intelligent navigation, cross-domain collaborative perception, and high mobility motion planning and control in dynamic open environment.
\end{IEEEbiography}

\begin{IEEEbiography}[{\includegraphics[width=1in,height=1.25in,clip,keepaspectratio]{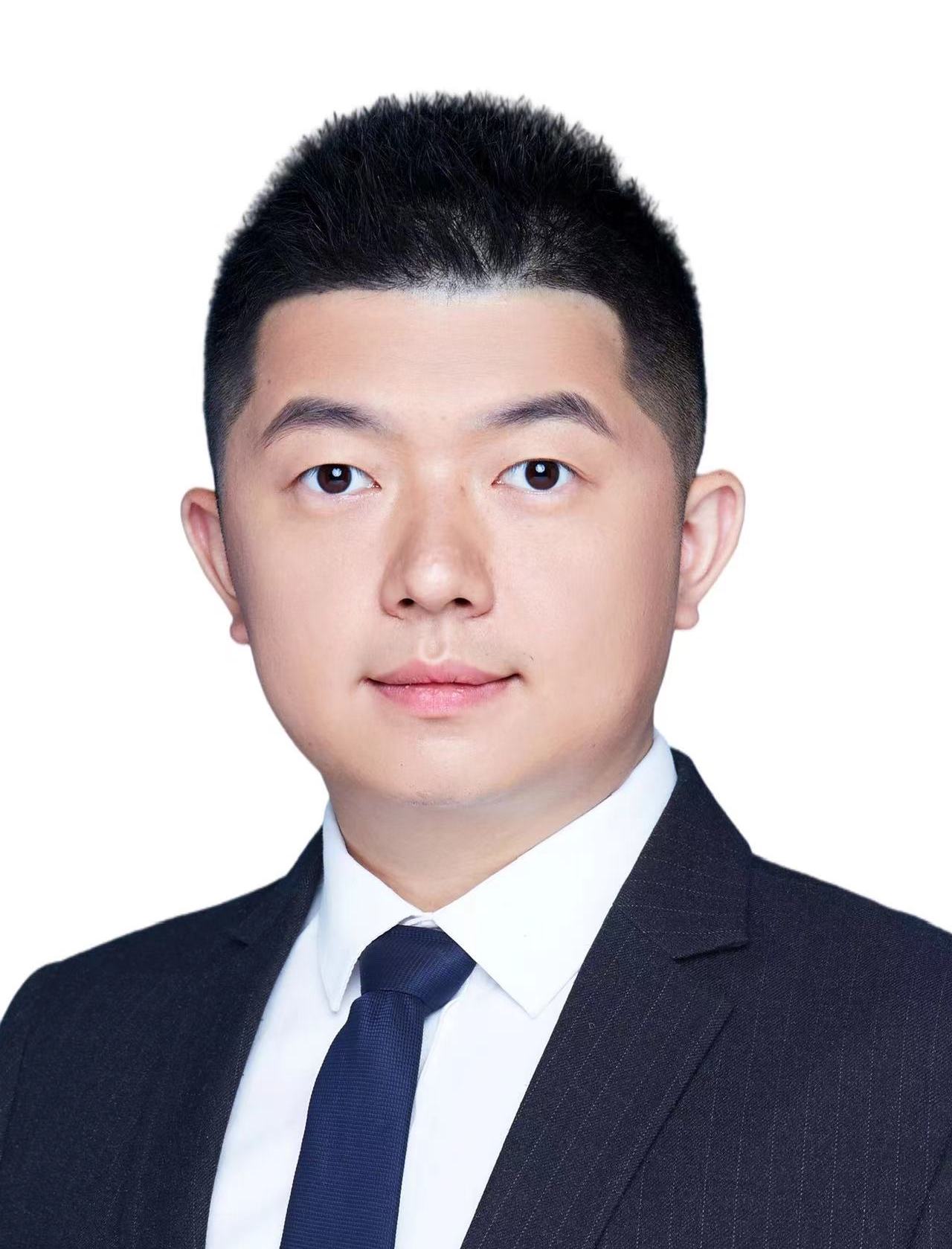}}]{Yufeng Yue}
(Member, IEEE) received the B.Eng. degree in automation from the Beijing Institute of Technology, Beijing, China, in 2014, and the Ph.D. degree in electrical and electronic engineering from Nanyang Technological University, Singapore, in 2019. 

He is currently a Professor with School of Automation, Beijing Institute of Technology. His research interests
include perception, mapping and navigation for autonomous robotics. He
has authored a book in Springer, and more than 60 journal/conference papers, including IEEE TNNLS/TCSVT/TMech/TIE, and NeurIPS/CVPR/ICRA/IROS.  He serves as an Associate Editor for  IEEE RAL/ICRA/IROS.
 \end{IEEEbiography}

\end{document}